%% file: main_aaai2027.tex
\documentclass[letterpaper]{article} 
\usepackage{aaai2027} 
\usepackage[hyphens]{url} 
\usepackage{graphicx} 
\usepackage{natbib} 
\usepackage{caption} 
\usepackage{subcaption}
\usepackage{amsmath}
\usepackage{amssymb}
\usepackage{booktabs}
\usepackage{array}
\usepackage{tikz}
\usepackage{pgfplots}
\pgfplotsset{compat=1.18}
\usetikzlibrary{patterns}
\usepackage{multirow}
\newcolumntype{C}[1]{>{\centering\arraybackslash}p{#1}}

\title{From Images to Grounded Edits: A Real-World Benchmark for Chinese Grammatical Error Correction in Handwritten Essays}

\author{
    Weiru Zhang\textsuperscript{\rm 1, \rm 2},
    Yuzhong Hong\textsuperscript{\rm 2},
    Junwei Bao\textsuperscript{\rm 2},
    Lemao Liu\textsuperscript{\rm 1},
    Dejing Dou\textsuperscript{\rm 1}
}
\affiliations{
    \textsuperscript{\rm 1}School of Computer Science, Fudan University\\
    \textsuperscript{\rm 2}Zuoyebang Education Technology, Inc.
}

\begin{document}
\maketitle

\begin{abstract}
Chinese Grammatical Error Correction (CGEC) is typically formulated as a \emph{text-to-text} problem. While suitable for digital texts, this formulation falls short in real-world K-12 education, where AI-assisted corrections are expected directly on essay images—which are primarily handwritten and exhibit many irregularities. To bridge this gap, we introduce Grounded Chinese Grammatical Error Correction (G-CGEC), an \emph{image-to-grounded-edit} task requiring systems to predict both corrections and their positions on essay images. We propose G-CGEC-Eval, a benchmark comprising real-world images of K–12 Chinese essays with fine-grained annotations. For evaluation, we adapt IoU-based metrics from object detection and propose a more tolerant alternative based on center-distance matching. Using this benchmark, we evaluate open-source and proprietary vision-language models under a direct single-pass setting, as well as settings with auxiliary components—OCR and character grounding—coordinated via pipelining or agentic harnessing. Results show that even top-tier proprietary models perform surprisingly poorly in the single-pass setting, while pipelining and agentic harnessing yield meaningful improvements. Notably, an error-type-based multi-agent workflow achieves the best performance, establishing it as the strongest baseline for future G-CGEC research. Through ablations and fine-grained analysis, we find the task remains far from solved, with linguistic error discovery and character-level grounding as the primary bottlenecks. This underscores the need for models that better integrate visual perception with linguistic reasoning and visual grounding.
\end{abstract}

\section{Introduction}

\begin{figure}[!t]
    \centering
    \includegraphics[width=1.0\columnwidth]{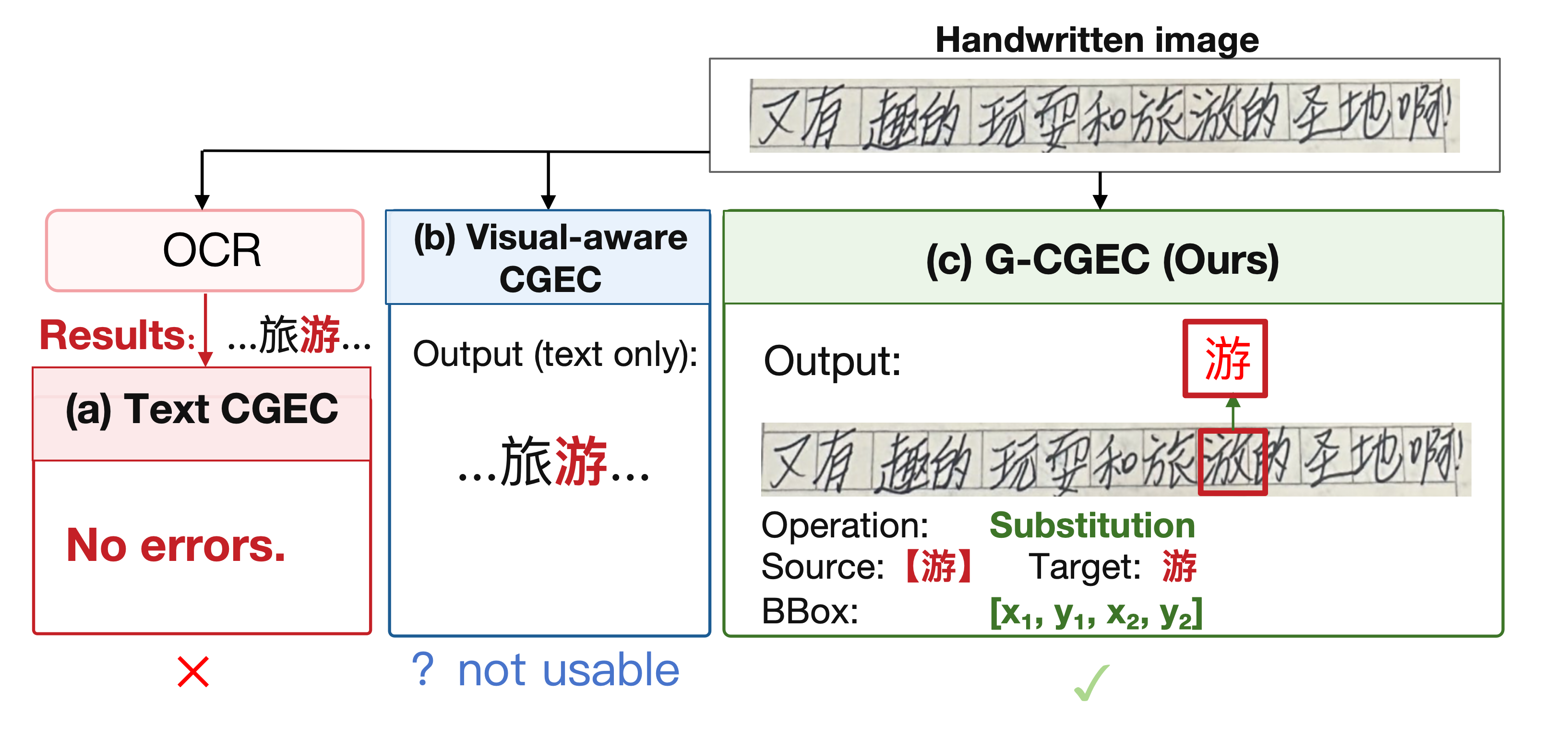} 
    \caption{When the input is in the image format, our image-to-grounded-edit formulation of the error correction task offers intuitive advantages: giving a system complete information as opposed to lossy OCR transcription; providing practical usefulness (e.g., for K-12 education).}
    \label{fig:ocr_error_propagation}
\end{figure}

Chinese Grammatical Error Correction (CGEC) has predominantly relied on text-based models \citep{Zhao2018OverviewOT, Zhang2022MuCGECAM}. However, these models struggle to meet the complex demands of handwritten essay correction in K-12 education. In real-world pedagogical scenarios, essays are often characterized by illegible handwriting, phonetic and visual character confusions, and irregular layouts, which impose a heavy cognitive burden on teachers. The performance of a naive ``OCR + Correction'' pipeline is highly dependent on recognition accuracy. As shown in Figure \ref{fig:ocr_error_propagation}(a), a failure of the OCR module unavoidably forces the correction module to process text that is inconsistent with the source image. For general NLP tasks, this error cascading effect of OCR is also a known issue \citep{vanStrien2020AssessingTI, Zhang2024OCRHR}.

 
This cascading effect means that purely text-based CGEC benchmarks cannot accurately capture end-to-end correction performance when the input is only available in the image format. To address this gap, recent CGEC research has begun incorporating visual features. However, existing visual-aware datasets still exhibit significant limitations in diversity, text density, and task paradigms. For example, VisCGEC \citep{wang2025viscgec} uses data from ideal settings, lacking the ``in-the-wild'' noise found in reality, such as uneven lighting, shadows, and perspective distortions. ReViCo \citep{Liang2025VisionLM} focuses on real-world scenarios, but the images are mainly photographs of signs, cards, and tags, containing only sparse text segments. Moreover, Visual-C3 \citep{Li2023TowardsRW} targets only Chinese Spell Checking (CSC), failing to cover multifaceted grammatical errors. Most crucially, although these studies integrate visual modalities, vision remains merely an auxiliary component for text-to-text correction. None of them require models to perform explicit visual grounding to pinpoint the location of errors on the original image.

Motivated by these observations, we introduce and study a more practically meaningful and challenging task: Grounded Chinese Grammatical Error Correction (G-CGEC), which requires a system to predict corrections directly on essay images. Our contributions are as follows:



\begin{itemize}
    \item We propose G-CGEC-Eval, a real-world benchmark dataset with fine-grained annotations of error corrections and their visual grounding. Collected from K--12 educational settings, it addresses a notable lack of visually diverse and textually dense handwritten essay datasets.

    \item We propose a visually-aware evaluation protocol that augments traditional text-based CGEC metrics. Specifically, we adapt IoU-based metrics from object detection and introduce a more tolerant center-distance matching strategy, which is particularly useful for assessing weak or intermediate-performing models.

    \item We benchmark representative open-source and proprietary VLMs, establishing strong baselines and identifying critical challenges in handling dense handwriting and complex semantic errors. This systematic analysis provides a foundation and clear directions for future research in multimodal CGEC.
\end{itemize}

\section{Related Work}

\subsection{Text-based CGEC Datasets}
Existing datasets for Chinese Grammatical Error Correction (CGEC) can be broadly categorized based on the types of errors they target, ranging from character-level spelling mistakes to sentence-level grammatical errors.

As a fundamental subtask of CGEC, Chinese Spelling Correction (CSC) focuses primarily on detecting and correcting erroneous characters. The most widely used benchmarks in this area are the SIGHAN datasets (SIGHAN13, 14, 15) \citep{Wu2013ChineseSC, Yu2014OverviewOS, Tseng2015IntroductionTS}. These datasets are derived from essays written by teenage students and Chinese as a Foreign Language (CFL) learners, capturing spelling mistakes typical of the learning process. Beyond open-domain texts, recent research has also expanded CSC into specific domains. For instance, MCSCSet \citep{Jiang2022MCSCSetAS} focuses on medical texts and ECSpell \citep{Lv2022GeneralAD} covers law and official documents.

Research on CGEC has evolved to address specific speaker profiles beyond simple spelling issues. Benchmarks targeting CFL learners, such as NLPCC \citep{Zhao2018OverviewOT}, CGED \citep{Rao2020OverviewON}, and MuCGEC \citep{Zhang2022MuCGECAM}, capture interlanguage errors. Conversely, CTC \citep{Zhao2022OverviewOC} and NaCGEC \citep{Ma2022LinguisticRC} focus on native-speaker errors. More recently, datasets such as FCGEC \citep{Xu2022FCGECFC} and FlaCGEC \citep{Du2023FlaCGECAC} introduced fine-grained annotations for precise error diagnostics. Despite this progress, existing CGEC benchmarks share a critical limitation: they rely solely on the textual modality.

\subsection{Visual-aware CGEC Datasets}
To leverage multimodal information, researchers have begun exploring visual aids for error correction. Visual-C3 \citep{Li2023TowardsRW} was proposed for visual Chinese Spelling Correction; however, it focuses exclusively on spelling errors rather than broader grammatical errors. VisCGEC \citep{wang2025viscgec} provides a dataset for visual Chinese grammatical error correction collected from CFL learners' handwritten texts, but is limited to sentence-level data on uniform grid paper backgrounds. ReViCo \citep{Liang2025VisionLM} features broader real-world scenarios, but only covers images with sparse text segments on signs, cards, and tags.

In contrast, our dataset encompasses a wide variety of complex grammatical issues. Furthermore, it features realistic background noise, dense texts with complete discourse-level contexts, and error corrections annotated with bounding boxes for localization and fine-grained labels for categorization.

\subsection{Evaluation Metrics for CGEC and Visual Grounding}
Due to the traditional text-to-text formulation of the grammatical error correction (GEC) task, evaluation metrics are purely text-based, including edit-based and fluency-based ones \cite{qiu2026chinesegrammaticalerrorcorrection}. Previous CGEC studies mainly adopt edit-based metrics with Chinese adaptations. The most general metric is the MaxMatch ($\text{M}^2$) scorer \citep{dahlmeier-ng-2012-better}, which aligns system edits to gold edits and reports precision, recall, and $\text{F}_{0.5}$. Because $\text{M}^2$'s word-segmentation dependence causes unstable alignments in Chinese, the field moved to ChERRANT, an extension of the English GEC evaluation tool ERRANT \citep{bryant-etal-2017-automatic} adapted for Chinese, introduced with the MuCGEC benchmark \citep{Zhang2022MuCGECAM}. ChERRANT produces linguistically-informed, rule-based edit alignments and error-type classifications (e.g., word order, redundant/missing/selection errors, punctuation) before computing the same precision/recall/$\text{F}_{0.5}$ scores at either the character or word level.

Considering the multimodal nature of the G-CGEC task, however, metrics from the computer vision domain for evaluating visual grounding accuracy are also relevant. The most relevant are Intersection over Union (IoU) \cite{everingham2010pascal,Zhang2013ObjectCD,Russakovsky2014ImageNetLS} and the precision/recall/F scores derived from it. These are standard in object detection for measuring the similarity between predicted and ground-truth bounding boxes, and are invariant to image translation, uniform scaling, 90\textdegree{}-multiple rotations, and reflections.

As detailed in later sections, our evaluation protocol \emph{combines} metrics from both fields, adapted and extended for the G-CGEC task.


\section{G-CGEC-Eval Benchmark}
\subsection{Task Formulation}
\label{sec:task_formulation}
We formally define the Grounded CGEC (G-CGEC) task for our benchmark as follows:
Given a sequence of input images $\mathcal{I}$ containing an essay, the expected output is a set of error correction edits: $E = \{e_1, \dots, e_n\}$. We define each edit $e_i$ by the following properties:
\begin{itemize}
    \item \textbf{Operation type} $\text{Op}[e_i]$: its valid value is one of $\{\text{Insertion, Deletion, Substitution}\}$.
    \item \textbf{Source} $\text{Src}[e_i]$ and \textbf{Target} $\text{Tgt}[e_i]$: the contents before and after the correction, respectively. For substitution operations, both are expected to be non-empty strings. For insertion operations, the source is expected to be an empty string and the target a non-empty string. For deletion operations, the opposite holds.
    \item \textbf{Image index} $\text{Idx}[e_i]\in \{0, \dots, |\mathcal{I}|-1\}$ and \textbf{Bounding box} $\text{Bbx}[e_i] = [x_{\min}, y_{\min}, x_{\max}, y_{\max}]$: the index of the image containing the error, and the coordinates of the top-left and bottom-right corners of the error region. For substitution and deletion operations, each box bounds the erroneous characters. For insertion operations, the box is placed at the intended insertion anchor, large enough to cover the space between the surrounding characters.
\end{itemize}

Note that we do not require the prediction of error type or reasons for each correction. This streamlines evaluation without compromising the core principle of the task: making correction edits on images. However, since we explicitly require the visual grounding outputs, the task is indeed more distinctly challenging and practically meaningful than previous benchmarks (e.g., ReViCo) that include the vision modality only as additional evidence for text-to-text correction.

\subsection{Multi-Reference Evaluation Protocol}
\label{sec:eval_design}

To comprehensively assess the model's performance on the Grounded CGEC task, we employ an evaluation protocol that considers both visual grounding accuracy and grammatical correction correctness. For clarity, we denote a ground truth edit as $e_i^{\text{true}}\in E^{\text{true}}$, and a predicted edit by a system in question as $e_j^{\text{pred}}\in E^{\text{pred}}$, where $E^{\text{pred}}$ represents the complete set of predicted edits, and $E^{\text{true}}$ the complete ground-truth set that \emph{best matches} $E^{\text{pred}}$ (selected as described in the following subsections).

Following the edit-based evaluation tradition in CGEC \cite{qiu2026chinesegrammaticalerrorcorrection}, we find the \emph{true positives} (TP) and \emph{false positives} (FP) in $E^{\text{pred}}$, and \emph{false negatives} (FN) in $E^{\text{true}}$. Then, we use these categorizations to compute precision (P), recall (R), and F1 using their standard definitions:
\begin{equation}
    \text{P} = \frac{\#\text{TP}}{\#\text{TP} + \#\text{FP}}
\end{equation}

\begin{equation}
    \text{R} = \frac{\#\text{TP}}{\#\text{TP} + \#\text{FN}}
\end{equation}

\begin{equation}
    \text{F}_1 = 2 \cdot \frac{\text{P} \cdot \text{R}}{\text{P} + \text{R}}
\end{equation}

We elaborate on the metric computation details in the following subsections.

\subsubsection{Categorization of TP, FP and FN} The categorization is based on both visual and textual matching between predicted edits and ground-truth edits. Specifically, we call $(e_i^{\text{true}}, e_j^{\text{pred}})\in E^{\text{true}}\times E^{\text{pred}}$ a \emph{textual match} if and only if $\text{Op}[e_i^{\text{true}}]=\text{Op}[e_j^{\text{pred}}]$, $\text{Src}[e_i^{\text{true}}]=\text{Src}[e_j^{\text{pred}}]$, and $\text{Tgt}[e_i^{\text{true}}]=\text{Tgt}[e_j^{\text{pred}}]$, i.e., an exact match of all textual contents. For a \emph{visual match}, however, an exact match is too strict to be practically meaningful. Instead, we only require $\text{Idx}[e_i^{\text{true}}]=\text{Idx}[e_j^{\text{pred}}]$ and their bounding-box similarity to exceed a given threshold. As commonly practiced in object detection \cite{everingham2010pascal}, we use IoU as one of our similarity measures, defined as follows: 
\begin{equation}
\text{\small$\text{IoU}(e_i^{\text{true}}, e_j^{\text{pred}}) = \frac{\text{Area}(\text{Bbx}[e_i^{\text{true}}] \cap \text{Bbx}[e_j^{\text{pred}}])}{\text{Area}(\text{Bbx}[e_i^{\text{true}}] \cup \text{Bbx}[e_j^{\text{pred}}])}$}
\end{equation}
While a cutoff on the IoU measure is less strict than an exact match, we find it is not tolerant enough. Even top-tier proprietary VLMs may predict a bounding box with very low or zero overlap with the ground truth, yet the deviation is small enough to remain acceptable for error localization. Such predictions should be considered better than those that are substantially off, but a strict IoU cutoff treats both equivalently as non-matches. To address this, we propose an alternative similarity measure called \textbf{center-distance}, defined as the Euclidean distance between the centroids of the bounding boxes (denoted as the vector-valued function $\textbf{c}(\cdot)$). All coordinates are normalized to a $0$--$1000$ scale, making center-distance invariant to uniform image scaling, just like IoU:

\begin{equation}
\text{\small$\text{CD}(e_i^{\text{true}}, e_j^{\text{pred}})
= \Big|\textbf{c}\big(\text{Bbx}[e_i^{\text{true}}]\big)-\textbf{c}\big(\text{Bbx}[e_j^{\text{pred}}]\big)\Big|$}
\end{equation}

After identifying all pairs that are both textual and visual matches, we sort them in descending order of IoU (or ascending order of center distance). We then traverse the sorted pairs to assign true positives greedily: for each pair $(e_i^{\text{true}}, e_j^{\text{pred}})$, we designate $e_j^{\text{pred}}$ as a TP and mark $e_i^{\text{true}}$ as matched, provided that neither has already been assigned. After traversal, any unmatched $e_i^{\text{true}}$ is categorized as FN, and any unassigned $e_j^{\text{pred}}$ is categorized as FP.

\subsubsection{Selecting the Best Matching Reference}
Some text segments in our dataset admit multiple valid correction solutions (see Section~\ref{sec:dataset}). To handle this ambiguity during evaluation, we enumerate all complete essay-level ground-truth sets by taking the Cartesian product of the solutions across all multi-solution segments, while keeping single-solution segments fixed. For instance, if one segment has two valid solutions and another has three, we obtain $2 \times 3 = 6$ complete ground-truth sets, each containing exactly one solution for every annotated segment.

For a given system prediction, we independently compute \#TP, \#FP, \#FN, precision, recall, and $F_1$ against each complete ground-truth set using the matching procedure described above. We then select a single set by sequential tiebreaking: we first pick the set(s) with the highest $F_1$-score; if multiple sets tie, we pick the one(s) with the most true positives; if still tied, the one(s) with the fewest false negatives; if still tied, the one(s) with the fewest false positives; if still tied, the one with the smallest index in the enumerated set.

\subsection{Dataset Construction}
\label{sec:dataset}
\begin{figure*}[t]
    \centering
    \input{figures/error_distribution.tex}
    \caption{Error category distribution in the G-CGEC-Eval benchmark. Horizontal bars show fine-grained categories grouped by 7 coarse categories (bold labels, separated by gray lines). All bars within a group share the same color. Note that syntactic, discourse-level errors, and punctuation errors together account for over 80\% of all errors.}
    \label{fig:error_distribution}
\end{figure*}
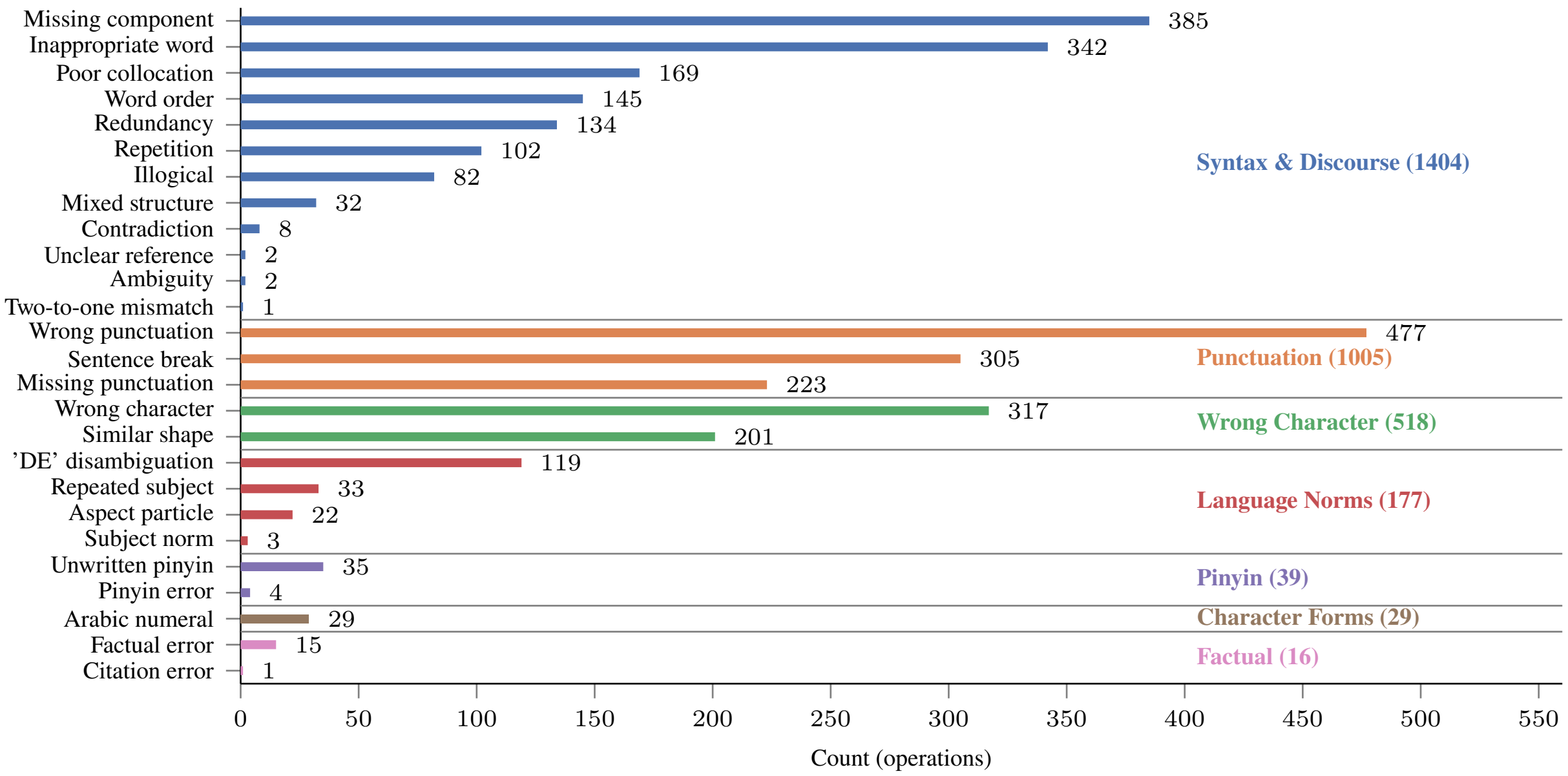

\begin{table}[t]
\centering
\small
\setlength{\tabcolsep}{6pt}
\begin{tabular}{l c}
\toprule
\textbf{Statistic} & \textbf{Value} \\
\midrule
Total essays & 150 \\
Total images & 251 \\
Total corrected text segments & 2\,731 \\
Total edits (primary) & 2\,913 \\
Avg. edits per segment & 1.1 \\
Edits with insertion operations & 849 (29.1\%) \\
Edits with deletion operations & 473 (16.2\%) \\
Edits with substitution operations & 1\,591 (54.6\%) \\
Segments with multiple solutions & 76 (2.8\%) \\
\bottomrule
\end{tabular}
\caption{Statistics of G-CGEC-Eval. Operations are counted from the primary solution of each segment.}
\label{tab:dataset_statistics}
\end{table}

The test dataset was collected to preserve the natural variation found in real-world K--12 handwritten essays. This section details the data collection and sampling procedures, the annotation process, and the final dataset statistics.

\subsubsection{Collection and Sampling}
First, raw data were collected through partnerships with several schools. We distributed digital questionnaires to students to obtain consent for the academic use of their handwritten essays, ultimately receiving authorization from over 6,000 students. Students were then asked to digitize their essays by taking photographs. We provided no specific instructions on how to capture the images, other than requiring them to contain a complete student essay. 

Second, we engaged professional data annotation service providers to review the collected data and strictly filter out unsuitable images based on the following criteria:
\begin{itemize}
    \item the image does not contain a legitimate Chinese essay (e.g., non-Chinese content or non-essay materials), because some students failed to follow our instructions;
    \item characters are unrecognizable due to extreme blurring, poor lighting, heavy shadows, severe perspective distortion, occlusion, or content cropped outside the frame;
    \item the essay is too short to provide meaningful discourse-level context for error correction (fewer than 20 characters).
\end{itemize}
After screening, we randomly sampled 150 essays to form the test dataset. Since our experiments focus on zero-shot and inference-time evaluation, no separate training or development sets are used.

\subsubsection{Annotation Process}
The annotation process consists of three stages: transcription, error correction, and bounding-box localization.

In the \textit{transcription} stage, professional annotators manually transcribe the handwritten text from each essay image. Annotators are instructed to preserve all original content faithfully, including grammatical errors, misspellings, and punctuation mistakes, without implicitly correcting anything. For non-existent characters, annotators guess valid characters that most closely resemble them visually and are contextually plausible, then enclose each guessed character in lenticular brackets, as shown in Figure~\ref{fig:ocr_error_propagation}. To ensure transcription quality, three annotators independently transcribe each essay, and disagreements are resolved through discussion or by a senior annotator.

In the \textit{error correction} stage, trained annotators with a background in Chinese language teaching identify grammatical, semantic, and spelling errors in the transcribed text. Each error is annotated with its operation type (insertion, deletion, or substitution), the source span, the corresponding corrected text, and the error category among 7 coarse types and 26 fine-grained subtypes. Some text segments can be corrected in multiple ways, each potentially involving multiple operations across different locations; annotators group such related operations into a single correction solution. For segments that admit multiple valid corrections, all mutually exclusive solutions are recorded. All other segments have at most one solution.

In the \textit{bounding-box localization} stage, annotators draw character-level bounding boxes around each operation in the original essay image. For substitution and deletion operations, the box covers the erroneous characters as they appear in the handwriting. For insertion operations, the box is placed at the intended insertion anchor, large enough to cover the space between the surrounding characters.

\subsubsection{Dataset Statistics}
Table~\ref{tab:dataset_statistics} and Figure~\ref{fig:error_distribution} summarize the composition of the dataset. The 150 essays span 251 images, yielding an average of 1.7 images per essay, which reflects the multi-page nature of handwritten essays. Across all essays, 2\,731 text segments requiring correction are annotated, corresponding to roughly 18 segments per essay. The low average of 1.1 operations per segment indicates that most corrections are localized and can be fixed with a single edit, though a non-trivial fraction require coordinated multi-operation solutions spanning several positions.

The error distribution reveals that syntax, discourse and punctuation errors together account for over 80\% of all operations, indicating that K--12 students frequently struggle with sentence structure, collocation, and proper punctuation usage. Wrong-character errors are the third most common category, arising from both visual similarity and phonetic confusion. Less frequent categories---language norms, pinyin, character forms, and factual errors---are still well-represented, ensuring that the benchmark tests a broad range of correction capabilities rather than focusing on a narrow subset of error types. The 76 segments with multiple valid solutions (2.8\%) highlight the inherent ambiguity of grammatical correction and motivate the multiple-reference evaluation protocol described in Section~\ref{sec:eval_design}.

\section{Experiments}
\subsection{Experimental Setup}
\subsubsection{Scope of Studied Systems}
The purpose of the experiments in this paper is to identify the challenges and common failure modes for our benchmark, and to establish strong baseline systems to facilitate future research on the G-CGEC task. Given the multimodal nature of G-CGEC, we employ Vision-Language Models (VLMs) as our foundational architecture, as they possess the joint visual and linguistic capabilities required to process essay images and generate grounded corrections. We evaluate three distinct inference schemes:

\begin{enumerate}
    \item \textbf{Single-pass Scheme}: In this end-to-end setting, the VLM is simply instructed to produce the required output given the input essay images, without any internal task decomposition or external auxiliary tools. It relies entirely on its inherent capabilities to generate grounded edits in a single step. This serves as the baseline for raw model capability. The complete prompt design and ablation results are reported in Appendix~\ref{sec:supp_prompt_sensitivity}.
    
    \item \textbf{Pipeline Scheme}: This modular approach decomposes the task into a rigid sequential flow. First, an OCR module provides a transcription of the essay. Second, the VLM performs textual error correction on the transcribed text. Finally, a character grounding module remaps the predicted textual edits back to the image coordinates. This scheme isolates the VLM's linguistic capabilities from the visual recognition burden.
    
    \item \textbf{Agentic Scheme}: This modular approach allows the VLM to dynamically coordinate with auxiliary tools (OCR and grounding models) through multi-step reasoning. Unlike the rigid pipeline, the agent decides when and how to invoke tools. We study two paradigms: (1) a \textit{single-agent} that handles all subtasks autonomously, and (2) a \textit{multi-agent} workflow where specialized sub-agents focus on either specific error types or operation types, and a master agent aggregates the results dynamically. All agentic experiments use the \textbf{OpenCode} framework. Detailed configurations are given in Appendix~\ref{sec:supp_implementation_details}.
\end{enumerate}

To ensure a fair comparison across schemes, we use the same collection of proprietary and open-source VLMs as representative backend models: GPT-5.5 (strongest proprietary), MiMo-V2.5 (large-scale open-source), and Qwen3.5-27B (small-scale open-source with moderate agentic capabilities). Whenever feasible, we set the temperature to $0.3$, the top-$p$ value to $1.0$, and the maximum generation length to $16{,}384$ tokens. Other decoding parameters follow the default settings of each model implementation. 

\subsubsection{OCR and Visual Grounding Modules}
We train the OCR module via SFT of Qwen3.5-35B-A3B on 26k image essays with annotated transcriptions. The transcription process is the same as the first stage of the annotation of our test dataset. 

The visual grounding module is also trained on top of Qwen3.5-35B-A3B using a rejection sampling-based SFT dataset. Concretely, we sample the raw visual grounding output using Gemini-3.1-pro-preview and select the samples that pass a simple verification test. The verification test consists of two steps: 
\begin{itemize}
    \item cropping the area around the predicted bounding boxes (i.e., \emph{zoom-in}) and then visualizing the box boundaries on the cropped images;
    \item using Qwen3.5-35B-A3B to recognize the text within the visualized boxes, and verifying the correctness of the box by checking whether the recognized text is consistent with the input text for visual grounding.
\end{itemize}

We strictly ensure that the benchmark dataset has no overlap with the training sets for these two modules. The intrinsic performance of the two modules and statistics of their training datasets are provided in Appendix~\ref{sec:supp_module_details}. In the main paper, we show their extrinsic impact through ablation studies.
\subsubsection{Evaluation Settings}
We primarily rely on the IoU-based matching strategy for visual matching in most of the experiments, with the IoU threshold set to 0.3 (denoted as IoU$@0.3$) by default unless explicitly specified otherwise. Whenever necessary, we further report performance under the center-distance-based matching strategy as a complement. 




\subsection{Main Results}
\begin{table}[t]
\centering
\small
\setlength{\tabcolsep}{1mm}
\begin{tabular}{l l r r r}
\toprule
\textbf{Scheme} & \textbf{Model} & \textbf{P (\%)} & \textbf{R (\%)} & $\mathbf{F_1}$\textbf{(\%)} \\
\midrule
\multirow{3}{*}{Single-pass} & Qwen3.5-27B & 1.86 & 0.55 & 0.85 \\
 & MiMo-V2.5 & 0.80 & 0.10 & 0.18 \\
 & GPT-5.5 & 0.53 & 0.14 & 0.22 \\
\midrule
\multirow{3}{*}{Pipeline} & Qwen3.5-27B & 18.69 & 10.55 & 13.48 \\
 & MiMo-V2.5 & 20.10 & 10.75 & 14.01 \\
 & GPT-5.5 & 16.32 & 9.97 & 12.37 \\
\midrule
\multirow{3}{*}{Agentic (E)} & Qwen3.5-27B & 11.94 & 3.33 & 5.20 \\
 & MiMo-V2.5 & 33.90 & 12.41 & 18.17 \\
 & GPT-5.5 & 33.11 & \textbf{17.42} & \textbf{22.83} \\
\midrule
Agentic (O) & GPT-5.5 & 27.45 & 9.96 & 14.61 \\
Agentic (S) & GPT-5.5 & \textbf{51.42} & 13.17 & 20.97 \\
\bottomrule
\end{tabular}
\caption{Main results across schemes and models. For Agentic schemes, (S), (E), (O) represents the \textbf{S}ingle-agent paradigm, the multi-agent paradigm with the \textbf{E}rror-type-based sub-agents, and the multi-agent paradigm with the \textbf{O}peration-type-based sub-agents, respectively.}
\label{tab:main_deployable_results}
\end{table}

\begin{table}[t]
\centering
\small
\setlength{\tabcolsep}{1mm}
\begin{tabular}{l l r}
\toprule
\textbf{Scheme} & \textbf{Model} & $\mathbf{F_1}$\textbf{(\%)} \\
\midrule
\multirow{3}{*}{VLM (Single-pass)} & Qwen3.5-27B & 0.85 \\
 & MiMo-V2.5 & 0.18 \\
 & GPT-5.5 & 0.22 \\
\midrule
\multirow{3}{*}{OCR + VLM} & Qwen3.5-27B & 1.13 \\
 & MiMo-V2.5 & 0.00 \\
 & GPT-5.5 & 0.38 \\
\midrule
\multirow{3}{*}{OCR + VLM + VG (Pipeline)} & Qwen3.5-27B & 13.48 \\
 & MiMo-V2.5 & 14.01 \\
 & GPT-5.5 & 12.37 \\
\bottomrule
\end{tabular}
\caption{Ablation experiments to study the individual impacts of the OCR module and the visual grounding (VG) module}
\label{tab:ablation_mods}
\end{table}

\begin{table}[t]
\centering
\small
\setlength{\tabcolsep}{1mm}

\begin{tabular}{l l r }
\toprule
\textbf{Scheme} & \textbf{Model} & $\mathbf{F_1}$\textbf{(\%)} \& $\Delta$ \\
\midrule
\multirow{3}{*}{OCR + VLM} & Qwen3.5-27B & 0.83 ($-$0.30) \\
 & MiMo-V2.5 & 0.06 ($+$0.06) \\
 & GPT-5.5 & 0.44 ($+$0.06) \\
\midrule
\multirow{3}{*}{Pipeline} & Qwen3.5-27B & 15.24 ($+$1.76) \\
 & MiMo-V2.5 & 16.37 ($+$2.36) \\
 & GPT-5.5 & 16.86 ($+$4.49) \\
\midrule
\multirow{3}{*}{Agentic (E)} & Qwen3.5-27B & 5.57 ($+$0.37) \\
 & MiMo-V2.5 & 19.32 ($+$1.15) \\
 & GPT-5.5 & 26.44 ($+$3.61) \\
\bottomrule
\end{tabular}
\caption{
Results of the oracle experiment to study the effect of OCR quality. Each cell reports the $F_1$ after replacing the OCR module results with the human-annotated transcriptions, with the absolute change shown in parentheses.
}
\label{tab:ocr_evidence_by_backend}
\end{table}

\subsubsection{Comparison Across System Schemes} Table~\ref{tab:main_deployable_results} presents the overall performance of all system schemes under all representative VLM backends. The results show that Grounded CGEC remains challenging even for current top-tier VLMs such as GPT-5.5 (proprietary) and MiMo-V2.5 (open-source). Direct single-pass inference yields the poorest performance across the board, whereas modular alternatives (pipelining and agentic harnessing) where text recognition and visual grounding are outsourced provide meaningful gains. The flexibility of the coordination of outsourced modules provided by the agentic setup is not always beneficial. For example, Qwen3.5-27B achieves worse results with agentic harnessing than with pipelining. However, for more powerful models, agentic harnessing generally leads to better results, with the strongest result achieved by GPT-5.5 under the error-type-based sub-agent setup. While agentic workflows incur higher inference costs due to multi-step reasoning and tool usage, their substantial performance gains justify their deployment for this challenging task.

\subsubsection{Comparison Across Agentic Workflows}
Table~\ref{tab:main_deployable_results} also compares single-agent inference, error-type-based sub-agents, and operation-type-based sub-agents under the same backend models. The error-type-based sub-agent workflow obtains the highest $F_1$, mainly by improving recall over the single-agent setting. This suggests that decomposing the task by error types helps the system discover more candidate errors. In contrast, operation-type-based sub-agents perform substantially worse.

\subsection{Ablation Studies}

\subsubsection{Individual Impact of OCR and Visual Grounding}
The pipeline scheme and the single-pass scheme serve as strict upper and lower bounds, representing the performance of VLMs with or without the two auxiliary modules. Therefore, to understand the individual impact of the two modules, we add an intermediate setting: a VLM is given the OCR results, but it predicts the outputs without the visual grounding module.

As shown in Table~\ref{tab:ablation_mods}, the boost provided by visual grounding alone is significant. This implies that current VLMs are weak at visual grounding for dense texts. This motivates the detailed failure mode analysis in Section~\ref{sec:failure_modes}. In contrast, OCR provides a negative or limited positive impact.

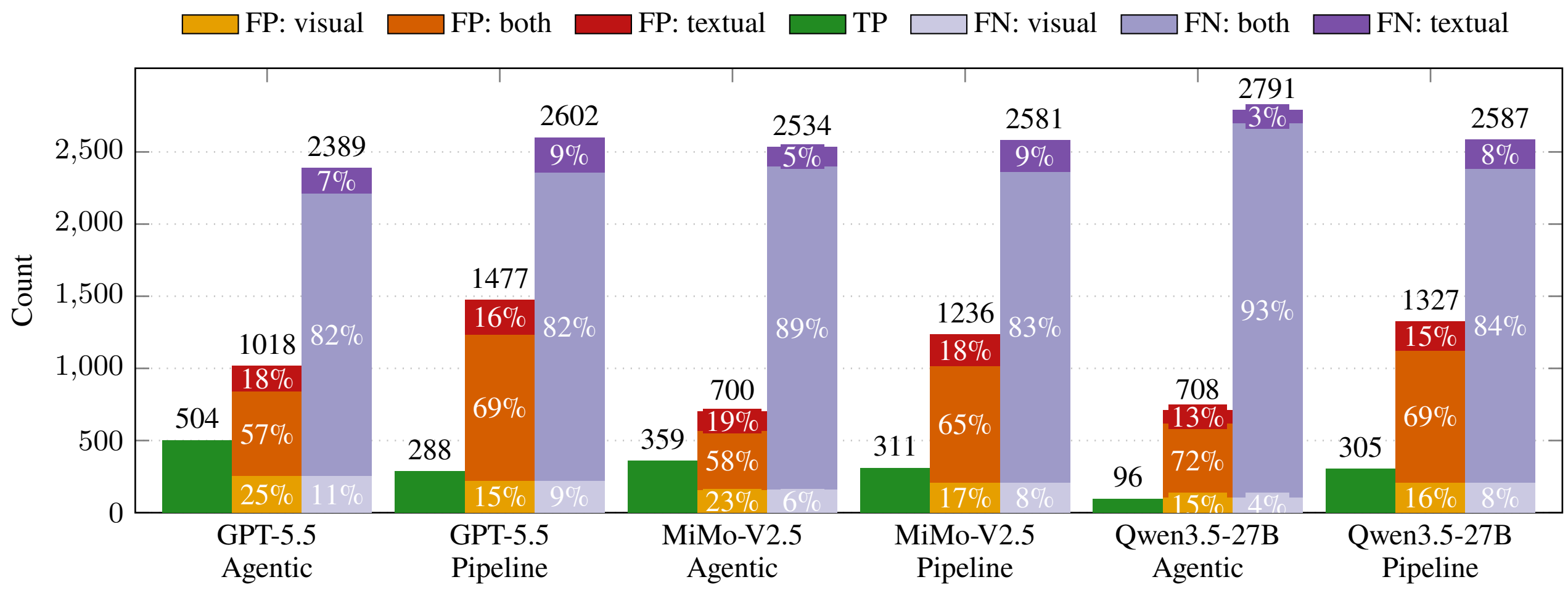
\begin{figure*}[!t]
    \centering
    \input{figures/failure_decomposition_counts}
    \caption{Fine-grained failure decomposition. Each configuration has three bars: TP, FP, and FN. Bar heights are raw counts. FP and FN bars are internally stacked by subtype, and in-bar percentages (rounded to integers) denote subtype proportions. In the legend, \emph{visual}, \emph{textual}, and \emph{both} indicate which kind of match failed.}
    \label{fig:failure_decomposition_counts}
\end{figure*}

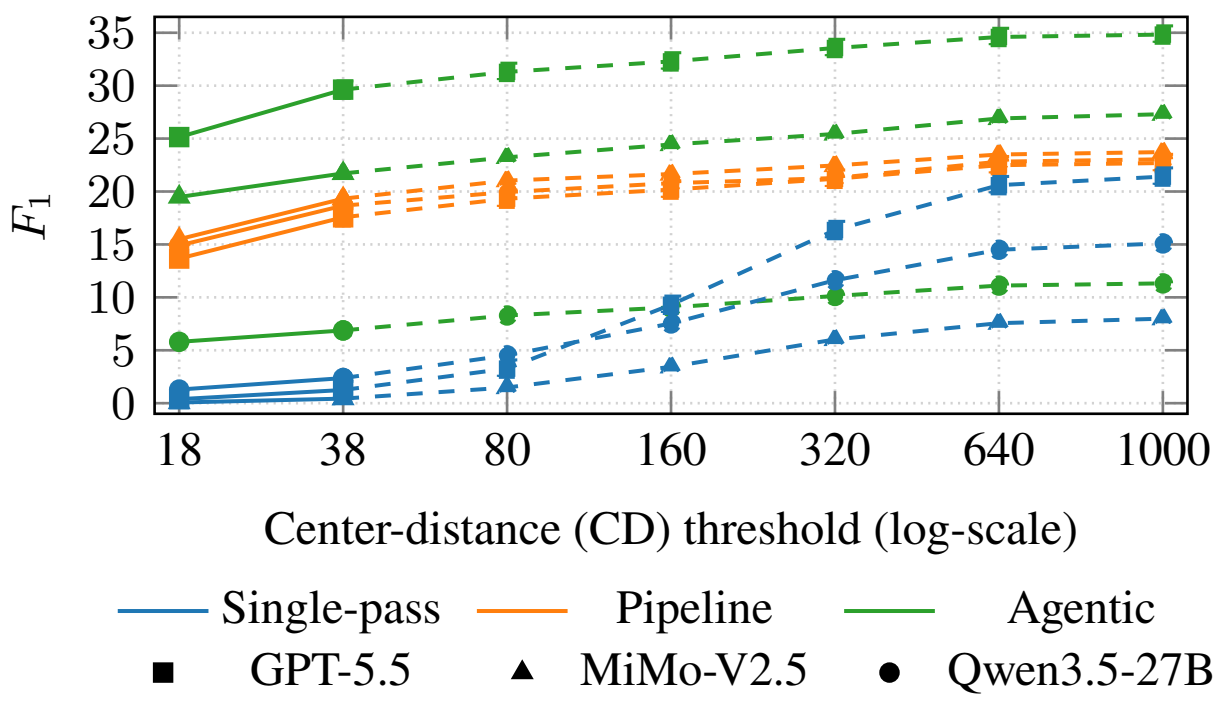
\begin{figure}[!t]
    \centering
    \input{figures/center_distance_curve}
    \caption{Center-distance-based analysis as a complement to Table \ref{tab:main_deployable_results}. Colors indicate inference paradigms (Single-pass, Pipeline, Agentic), while marker shapes distinguish individual systems. Line segments up to CD@$38$ are shown as solid lines and the remaining segments are shown as dashed lines.}
    \label{fig:center_distance_curve}
\end{figure}

\subsubsection{Impact of OCR Quality}
Given the previous results, it is reasonable to further study why the OCR module provides limited improvement. Therefore, we conduct an oracle experiment by replacing the OCR results with the human transcription data (an intermediate artifact from the annotation process of the benchmark dataset). As shown in Table~\ref{tab:ocr_evidence_by_backend}, perfect OCR quality (i.e., human transcription) can significantly boost performance. 
This verifies the intuition that the error cascading effect as we exemplify in Figure~\ref{fig:ocr_error_propagation} is non-trivial, and our evaluation protocol can capture its impact.
Thus, despite the more profound impact of visual grounding, we argue that OCR quality is also a significant factor affecting performance.

\subsection{Failure Mode Studies}
\label{sec:failure_modes}
\subsubsection{Center-Distance-Based Diagnostics}

As discussed in Section~\ref{sec:eval_design}, the IoU measure offers limited flexibility in controlling the strictness of visual grounding evaluation. The center-distance (CD) metric addresses this limitation by allowing us to assess performance across a wider range of spatial tolerances. At lower thresholds, center-distance aligns with the strictness of IoU-based metrics: for instance, CD$@18$ corresponds closely to IoU$@0.3$, and CD$@38$ to IoU$@0.01$ (see Appendix~\ref{sec:supp_iou_cd_mapping} for empirical evidence). At higher thresholds, center-distance extends beyond the practical range of IoU: by progressively relaxing the spatial constraint, it isolates the impact of textual correction ability from visual grounding precision.

As shown in Figure \ref{fig:center_distance_curve}, as the center-distance threshold is relaxed, most systems achieve higher $F_1$ scores. This trend indicates that many predictions are textually plausible but visually imprecise. However, even when the spatial constraint is effectively relaxed at CD$@1000$, we observe that $F_1$ does not exceed 40\%. This indicates that textual error correction itself remains highly challenging, even for the strongest baseline systems.

\subsubsection{Fine-grained Error Analysis}
\label{sec:fine_grained_error_analysis}

We decompose false positives and false negatives into those caused by textual mismatch alone, visual mismatch alone, or both occurring simultaneously. As illustrated in Figure~\ref{fig:failure_decomposition_counts}, textual and visual mismatches contribute roughly equally to the overall error rate. However, the proportion of cases where both failures co-occur is conspicuously dominant. This reveals that visual grounding and linguistic reasoning are deeply coupled: a deficiency in one modality frequently exacerbates or aligns with a deficiency in the other, underscoring the need for models that tightly integrate spatial localization with contextual grammatical understanding.

\section{Conclusion}
In this paper, we introduced the task of G-CGEC and presented G-CGEC-Eval, the first benchmark for correcting errors directly on real-world essay images. We proposed a unified evaluation protocol combining exact textual matching with flexible spatial criteria (IoU and center-distance). Our experiments reveal that while direct VLM inference performs poorly, modular approaches that outsource OCR and visual grounding yield substantial gains, with an error-type-based multi-agent workflow setting the strongest baseline. Fine-grained analysis identifies character-level grounding and linguistic error discovery as the primary bottlenecks. Currently, our study focuses on Chinese K-12 essays and evaluates models without task-specific fine-tuning. Furthermore, the strong performance of agentic workflows comes at the cost of higher inference latency, highlighting a need for more efficient reasoning strategies. We hope G-CGEC-Eval serves as a rigorous testbed for advancing multimodal error correction systems that better integrate visual perception with linguistic reasoning. 

\clearpage
\bibliography{custom}
\clearpage
\appendix

\section{Complete IoU@$0.3$ Results}
\label{sec:supp_complete_iou03_results}
This section complements the main paper by reporting IoU@$0.3$ results for a broader range of VLM backends. Table~\ref{tab:supp_complete_iou03_results} presents results for deployable systems (using the OCR module) across the single-pass, pipeline, and agentic schemes. Table~\ref{tab:supp_human_ocr_results} provides the corresponding oracle results using human-annotated transcriptions.

\section{Empirical Mapping between IoU and Center Distance}
\label{sec:supp_iou_cd_mapping}

The main paper uses IoU as the primary visual matching criterion and center distance (CD) as a complementary diagnostic. Since CD operates on a normalized 0\textendash1000 coordinate scale, raw threshold values are not immediately intuitive. We therefore establish concrete reference points by measuring how closely CD-based pair selection aligns with IoU-based pair selection across systems. This mapping allows readers to interpret CD thresholds in terms of the more familiar IoU scale.

\subsection{Method}
To align this analysis with the main comparison, we use the Single-pass and Agentic (E) predictions from Qwen3.5-27B, MiMo-V2.5, and GPT-5.5, yielding six systems in total. For each system, we collect the complete predictions across all 150 benchmark essays. Within each essay image, we enumerate all textually matched edit prediction pairs $(e_i^{\text{true}}, e_j^{\text{pred}})$ and compute both IoU and CD.

We treat IoU-matched edit pairs as the reference and measure how well CD matching recovers them. For an IoU threshold $\tau$ and a CD threshold $r$, an edit pair is an IoU match if $\text{IoU}(e_i^{\text{true}}, e_j^{\text{pred}}) \geq \tau$ and a CD match if $\text{CD}(e_i^{\text{true}}, e_j^{\text{pred}}) \leq r$. We compute the precision and recall of CD matching against IoU matching:
\begin{equation}
P_\tau(r) = \frac{|\{(e_i^{\text{true}}, e_j^{\text{pred}}) : \text{CD} \leq r \land \text{IoU} \geq \tau\}|}{|\{(e_i^{\text{true}}, e_j^{\text{pred}}) : \text{CD} \leq r\}|},
\end{equation}
\begin{equation}
R_\tau(r) = \frac{|\{(e_i^{\text{true}}, e_j^{\text{pred}}) : \text{CD} \leq r \land \text{IoU} \geq \tau\}|}{|\{(e_i^{\text{true}}, e_j^{\text{pred}}) : \text{IoU} \geq \tau\}|}.
\end{equation}
\begin{equation}
F_{1,\tau}(r)
=
2\cdot
\frac{
P_{\tau}(r)R_{\tau}(r)
}{
P_{\tau}(r)+R_{\tau}(r)
}.
\end{equation}
For each system and each $\tau$, we sweep $r$ from $0$ to $100$ (step 1) and find the $r$ that maximizes $F_1$. We report both the distribution of per-system optimal $r$ values and the aggregate median P/R/$F_1$ curves across systems.

\begin{table}[!t]
\centering
\small
\setlength{\tabcolsep}{3pt}
\begin{tabular}{l l r r r}
\toprule
\textbf{Scheme} & \textbf{Model} & \textbf{P (\%)} & \textbf{R (\%)} & $\mathbf{F_1}$\textbf{ (\%)} \\
\midrule
Single-pass & Gemini 3.1 Pro & 11.10 & 3.19 & 4.95 \\
 & Qwen3.5-397B-A17B & 6.04 & 1.70 & 2.65 \\
 & Qwen3.5-27B & 1.86 & 0.55 & 0.85 \\
 & GPT-5.5 & 0.53 & 0.14 & 0.22 \\
 & MiMo-V2.5 & 0.80 & 0.10 & 0.18 \\
\midrule
Pipeline & Gemini 3.1 Pro & 22.50 & 15.32 & 18.23 \\
 & MiMo-V2.5 & 20.10 & 10.75 & 14.01 \\
 & Qwen3.6-27B & 19.44 & 10.55 & 13.68 \\
 & Qwen3.5-27B & 18.69 & 10.55 & 13.48 \\
 & GPT-5.5 & 16.32 & 9.97 & 12.37 \\
\midrule
Agentic (E) & GPT-5.5 & 33.11 & 17.42 & \textbf{22.83} \\
 & MiMo-V2.5 & 33.90 & 12.41 & 18.17 \\
 & Qwen3.6-27B & 17.81 & 7.26 & 10.31 \\
 & Qwen3.5-27B & 11.94 & 3.33 & 5.20 \\
\midrule
Agentic (O) & GPT-5.5 & 27.45 & 9.96 & 14.61 \\
Agentic (S) & GPT-5.5 & \textbf{51.42} & 13.17 & 20.97 \\
\bottomrule
\end{tabular}
\caption{Deployable system results across a broader set of VLM backends. Agentic (S), (E), and (O) denote single-agent, error-type-based sub-agent, and operation-type-based sub-agent workflows, respectively.}
\label{tab:supp_complete_iou03_results}
\end{table}

\begin{table}[!t]
\centering
\small
\setlength{\tabcolsep}{3pt}
\begin{tabular}{l l l r r r}
\toprule
\textbf{Scheme} & \textbf{Model} & \textbf{P (\%)} & \textbf{R (\%)} & $\mathbf{F_1}$\textbf{ (\%)} \\
\midrule
Single-pass & Gemini 3.1 Pro & 16.00 & 4.43 & 6.94 \\
 & Qwen3.5-27B & 1.64 & 0.55 & 0.83 \\
 & GPT-5.5 & 1.04 & 0.28 & 0.44 \\
 & MiMo-V2.5 & 0.26 & 0.03 & 0.06 \\
\midrule
Pipeline & Gemini 3.1 Pro & 28.26 & 18.80 & 22.58 \\
 & GPT-5.5 & 18.86 & 15.24 & 16.86 \\
 & MiMo-V2.5 & 22.79 & 12.77 & 16.37 \\
 & Qwen3.6-27B & 20.99 & 12.01 & 15.28 \\
 & Qwen3.5-27B & 20.49 & 12.14 & 15.24 \\
\midrule
Agentic (E) & GPT-5.5 & 37.80 & 20.32 & \textbf{26.44} \\
 & MiMo-V2.5 & 36.00 & 13.20 & 19.32 \\
 & Qwen3.6-27B & 16.93 & 8.30 & 11.14 \\
 & Qwen3.5-27B & 10.89 & 3.74 & 5.57 \\
\midrule
Agentic (S) & GPT-5.5 & \textbf{57.01} & 14.93 & 23.66 \\
\bottomrule
\end{tabular}
\caption{Oracle results using human-annotated transcriptions across additional VLM backends. Agentic (S) and (E) denote single-agent and error-type-based sub-agent workflows, respectively.}
\label{tab:supp_human_ocr_results}
\end{table}

\subsection{Results}
Table~\ref{tab:supp_iou_cd_pair_mapping} summarizes the empirical mapping. For IoU@$0.3$, the median optimal distance across the six systems is $r=18$, and the aggregate median $F_1$ curve also peaks at $r=18$. For IoU@$0.01$, the median optimal distance is $r=36$, with the aggregate curve peaking at $r=38$. These values establish the reference points used in the main paper: CD@$18$ $\approx$ IoU@$0.3$ and CD@$38$ $\approx$ IoU@$0.01$. They provide a direct way to interpret center-distance thresholds on the familiar IoU scale.

\subsection{Robustness Check}
\begin{figure*}[t]
    \centering
    \begin{subfigure}{0.45\textwidth}
        \centering
        \includegraphics[width=\textwidth]{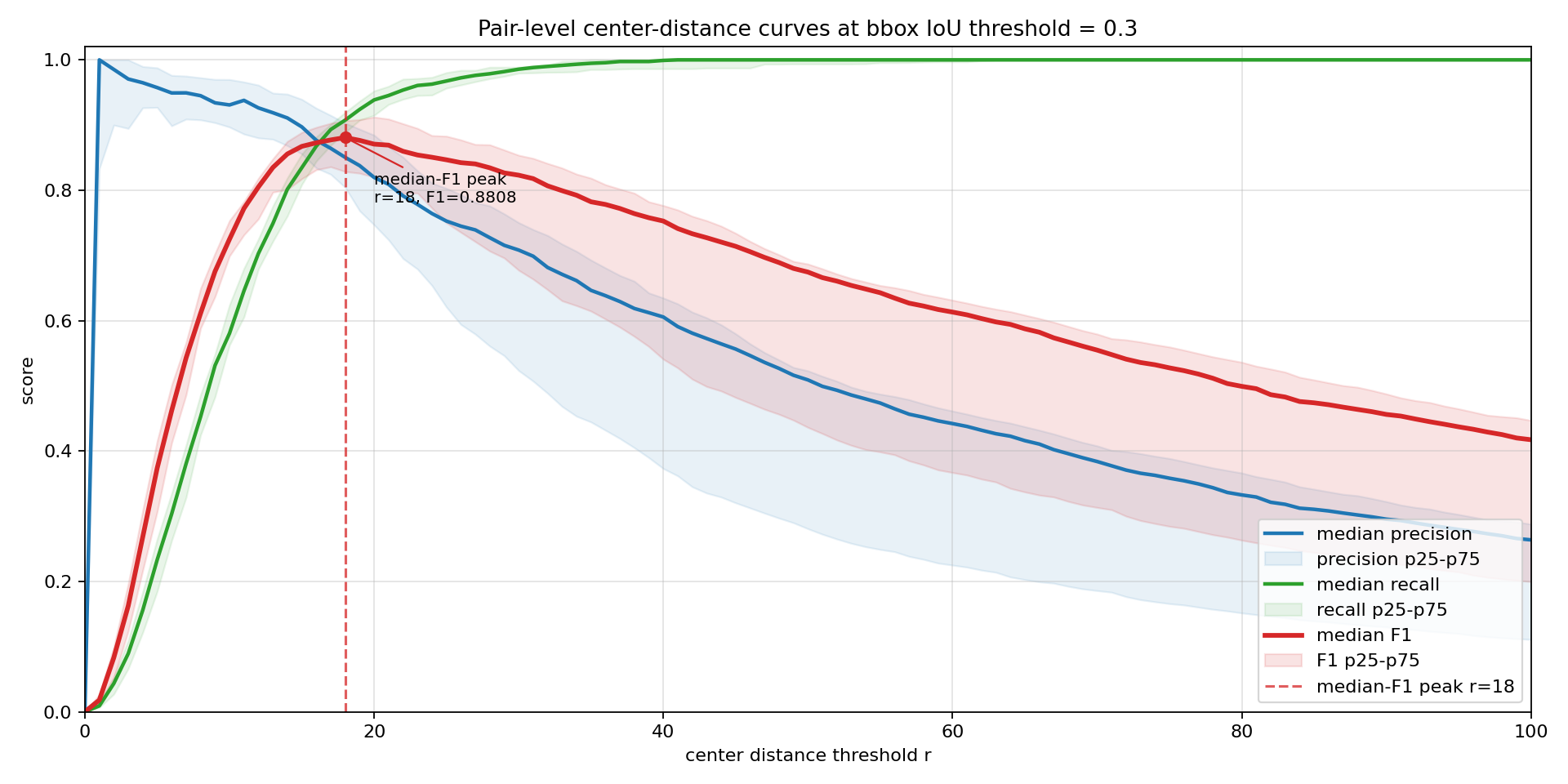}
        \caption{IoU threshold $0.3$ (peak at $r=18$).}
    \end{subfigure}
    \hfill
    \begin{subfigure}{0.45\textwidth}
        \centering
        \includegraphics[width=\textwidth]{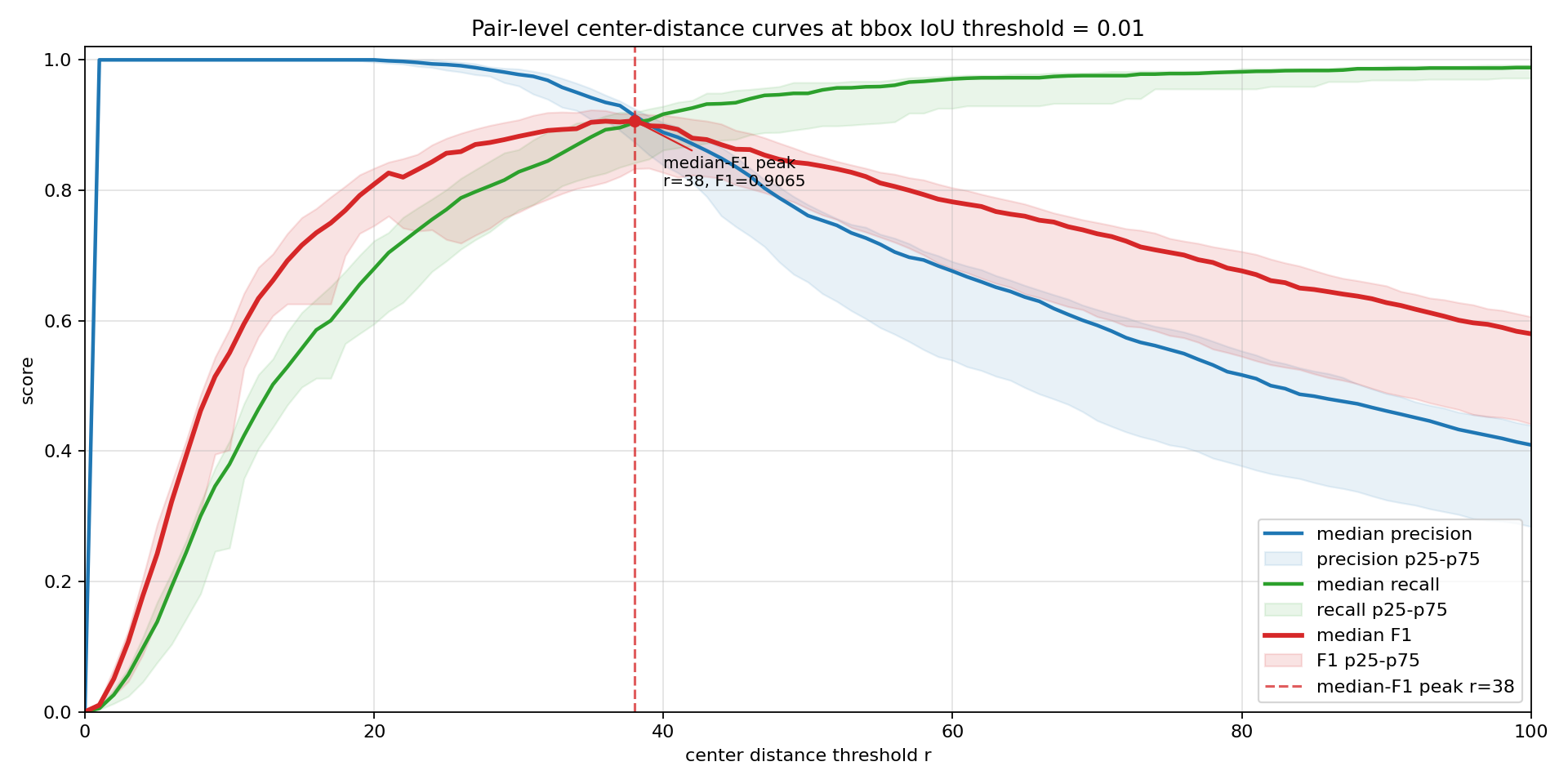}
        \caption{IoU threshold $0.01$ (peak at $r=38$).}
    \end{subfigure}
    \caption{Pair-level precision, recall, and $F_1$ of CD matching against IoU matching across the 37 system outputs. Shaded regions denote the 25\textendash75 percentile range across systems.}
    \label{fig:supp_pair_level_prf_curves}
\end{figure*}

To verify that these reference points generalize, we repeat the analysis on a broader pool of 37 system outputs. This set covers single-pass, pipeline, agentic configurations with varying OCR settings, spanning GPT-5.4/5.5, Gemini 3 Flash/3.1 Pro, Qwen3.5, Qwen3.6 Plus, MiMo-V2.5, Claude Opus 4.7, etc. As shown in Figure~\ref{fig:supp_pair_level_prf_curves}, the median $F_1$ peaks remain stable at $r=18$ for IoU@$0.3$ and $r=38$ for IoU@$0.01$. This confirms that the IoU--CD correspondence holds consistently across diverse models and inference paradigms.

\subsection{Instance-level illustration}
\begin{figure*}[t]
    \centering
    \includegraphics[width=0.98\textwidth]{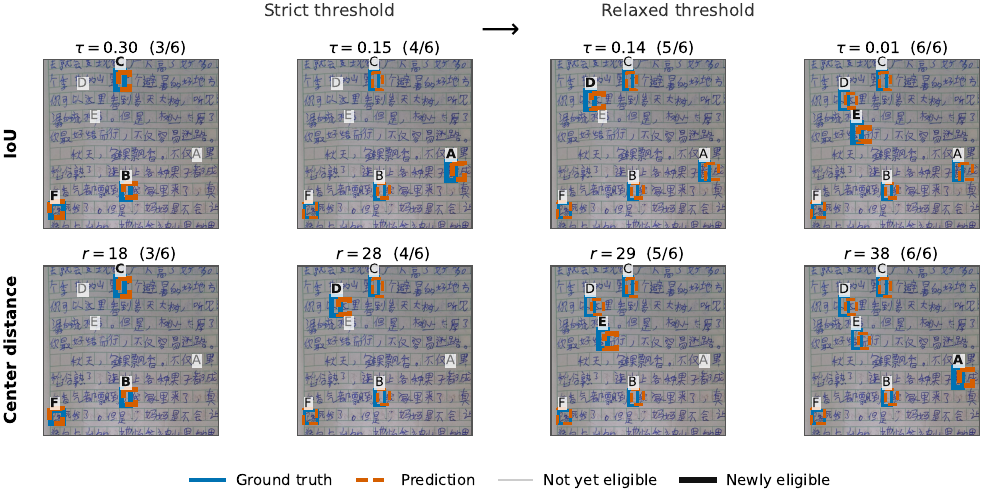}
    \caption{Instance-level illustration of the IoU--CD correspondence. Top row: progressively relaxing the IoU threshold. Bottom row: increasing the CD threshold $r$. Blue solid boxes are ground truth, orange dashed boxes are predictions. Gray boxes are not yet visually eligible; thicker outlines indicate pairs newly admitted at each threshold. Intermediate thresholds are case-specific breakpoints chosen for visualization.}
    \label{fig:spatial_threshold_case_study}
\end{figure*}

Figure~\ref{fig:spatial_threshold_case_study} visualizes the correspondence at the instance level, using predictions from the strongest system (GPT-5.5 with error-type-based sub-agents). We select pairs whose textual edits match exactly and vary only the visual threshold. At both reference points, IoU and CD admit the same instances: IoU@$0.3$ and CD@$18$ each admit three (Cases B, C, F), while IoU@$0.01$ and CD@$38$ each admit all six. Their intermediate trajectories differ---IoU relaxation first admits Case A, whereas CD relaxation first admits Case D---confirming that the criteria are not geometrically equivalent, but that the reference points capture a meaningful empirical alignment.

\begin{table*}[t]
\centering
\small
\setlength{\tabcolsep}{3pt}
\begin{tabular}{C{0.07\textwidth} C{0.14\textwidth} C{0.11\textwidth} C{0.11\textwidth} C{0.10\textwidth} C{0.09\textwidth} C{0.09\textwidth} C{0.09\textwidth}}
\toprule
\textbf{IoU threshold}
& \textbf{Per-system optimal-$r$ range}
& \textbf{Per-system optimal-$r$ median}
& \textbf{Per-system optimal-$r$ mean}
& \textbf{Median-curve peak $r$}
& \textbf{Median $P$ at peak}
& \textbf{Median $R$ at peak}
& \textbf{Median $F_1$ at peak} \\
\midrule
$0.3$  & $0$--$22$  & $18$ & $16.54$ & $18$ & $0.8504$ & $0.9075$ & $0.8808$ \\
$0.01$ & $19$--$42$ & $36$ & $34.38$ & $38$ & $0.9145$ & $0.9039$ & $0.9065$ \\
\bottomrule
\end{tabular}
\caption{Empirical pair-level mapping between IoU thresholds and CD thresholds. For each IoU threshold $\tau$, we report the distribution of per-system optimal $r$ (the $r$ that maximizes pair-level $F_1$) and the aggregate median P/R/$F_1$ at the peak.}
\label{tab:supp_iou_cd_pair_mapping}
\end{table*}

\section{Pass@$k$ Metric}
\label{sec:supp_passk}
The main paper evaluates system performance using single-shot Precision, Recall, and $F_1$. However, VLMs are inherently stochastic, and a model may possess the capability to identify an error but fail to produce the correct prediction deterministically in a single inference. Pass@$k$ addresses this by measuring the probability that at least one out of $k$ independently sampled inferences correctly identifies a ground-truth edit. This metric is particularly valuable for our prompt sensitivity analysis, as it reveals how reliably a prompt guides the model toward correct answers across multiple samples, rather than relying on a single stochastic outcome.

In G-CGEC, each ground-truth error $e_i^{\text{true}}$ is treated as an independent instance. For a single model inference, we check whether it contains a predicted edit $e_j^{\text{pred}}$ that matches $e_i^{\text{true}}$. We evaluate two matching criteria: visual-only match and full match, where the latter considers both textual and visual matches as defined in the main paper.
Given $n$ generated samples ($n \geq k$), let $c$ be the number of samples that contain at least one matching prediction for $e_i^{\text{true}}$. The pass@$k$ score for this instance is estimated as:
\begin{equation}
\text{pass}@k = 1 - \frac{\binom{n-c}{k}}{\binom{n}{k}}.
\end{equation}
The overall pass@$k$ is computed by averaging this value over all ground-truth edit instances.

\section{Prompt Sensitivity Analysis}
\label{sec:supp_prompt_sensitivity}

The single-pass scheme relies entirely on the prompt to guide the model. To ensure a fair comparison across models and paradigms in the main paper, we conduct a systematic ablation study to identify a prompt that performs well and consistently across VLMs, rather than one optimized for any particular model.

\subsection{Prompt Design Space}
We decompose the single-pass prompt into three dimensions salient for G-CGEC and construct seven variants by controlled ablation:

\begin{itemize}
    \item \textbf{Instruction complexity:} Full specification (role definition, recognition rules, annotation conventions, detailed constraints on punctuation, temporal expressions, special cases, and bounding-box format) versus condensed essentials.
    \item \textbf{Transcription output:} Whether the model outputs the full essay transcription before producing corrections.
    \item \textbf{Output format:} XML-style tags (e.g., \texttt{<TTL>}, \texttt{<PARA>}, \texttt{<point>}) versus JSON. For JSON, we further test a \textit{decoupled} variant that separates textual correction from bounding-box grounding into distinct output stages.
\end{itemize}

\subsection{Development Set for Prompt Selection}
To avoid overfitting the prompt to the specific models evaluated in the main paper, we use a separate development set of VLM backends for the prompt sensitivity analysis. These models are not included in the main experimental comparisons but share similar architectural characteristics and capabilities. By optimizing the prompt on this independent set, we ensure that the selected default prompt generalizes well across models rather than being tuned to the idiosyncrasies of a particular backend.

For this analysis, we evaluate all prompt variants on a 10-essay subset, sampling $n=32$ generations per backend and computing pass@$k$ for $k \in \{1,2,4,8,16\}$ under IoU@$0.3$. Table~\ref{tab:supp_passk_results} reports the pass@$k$ results for the selected default prompt (Variant 1). Pass@$k$ increases monotonically with the sampling budget for every backend, confirming that repeated sampling reliably uncovers additional correct predictions. The gap between visual-only and full-match scores is substantial: at $k=16$, the strongest visual-only score reaches 48.39\%, whereas the strongest full-match score is only 27.16\%. This indicates that while repeated sampling improves the chance of locating relevant regions, correct localization alone does not resolve the textual correction bottleneck. The backend ordering also shifts depending on the matching criterion: Gemini 3 Flash leads under visual-only pass@$16$, while Qwen3.5-397B-A17B leads under full-match pass@$16$.

\begin{table*}[t]
\centering
\small
\setlength{\tabcolsep}{2.7pt}
\begin{tabular}{@{}l r r r r r r r r r r@{}}
\toprule
& \multicolumn{5}{c}{\textbf{Visual-only pass@$k$ (\%)}} & \multicolumn{5}{c}{\textbf{Full-match pass@$k$ (\%)}} \\
\cmidrule(lr){2-6}\cmidrule(l){7-11}
\textbf{Model} & \textbf{@1} & \textbf{@2} & \textbf{@4} & \textbf{@8} & \textbf{@16} & \textbf{@1} & \textbf{@2} & \textbf{@4} & \textbf{@8} & \textbf{@16} \\
\midrule
Qwen3-VL-235B-A22B-Thinking & 3.97 & 6.72 & 10.32 & 14.56 & 19.45 & 0.38 & 0.73 & 1.35 & 2.34 & 3.70 \\
Qwen3-VL-30B-A3B-Instruct & 9.92 & 15.63 & 23.33 & 33.28 & 43.94 & 1.48 & 2.83 & 5.19 & 8.87 & 13.46 \\
Qwen3-VL-32B-Instruct & 5.53 & 9.42 & 14.60 & 20.86 & 28.91 & 1.31 & 2.33 & 3.80 & 5.48 & 7.43 \\
Qwen3-VL-32B-Thinking & 1.69 & 3.20 & 5.78 & 9.62 & 14.28 & 0.04 & 0.08 & 0.17 & 0.34 & 0.68 \\
Qwen3-VL-8B-Instruct & 2.32 & 3.35 & 4.95 & 7.33 & 10.34 & 0.00 & 0.00 & 0.00 & 0.00 & 0.00 \\
Qwen3-VL-8B-Thinking & 0.21 & 0.42 & 0.83 & 1.61 & 3.05 & 0.00 & 0.00 & 0.00 & 0.00 & 0.00 \\
Gemini 3 Flash & 10.52 & 18.01 & 27.91 & 38.52 & \textbf{48.39} & 5.41 & 9.48 & 15.07 & 20.90 & 25.01 \\
Qwen3.5-35B-A3B & 3.46 & 6.50 & 11.54 & 18.69 & 27.28 & 1.31 & 2.42 & 4.18 & 6.42 & 8.48 \\
Qwen3.5-397B-A17B & 9.59 & 16.01 & 24.46 & 34.56 & 46.20 & 4.05 & 7.04 & 11.66 & 18.37 & \textbf{27.16} \\
Qwen3.5-Flash & 3.00 & 5.62 & 10.02 & 16.60 & 25.13 & 1.10 & 2.01 & 3.41 & 5.23 & 7.20 \\
\bottomrule
\end{tabular}
\caption{Pass@$k$ results for the selected default prompt (Variant 1) at IoU@$0.3$. Each backend has 32 sampled generations on the same 10-essay subset. Scores are percentages; Visual-only requires only spatial matching, while Full-match additionally requires the textual edit to match exactly.}
\label{tab:supp_passk_results}
\end{table*}

\begin{table*}[t]
\centering
\small
\setlength{\tabcolsep}{2pt}
\begin{tabular}{l c c c | c c c c}
\toprule
\multirow{2}{*}{\textbf{Variant}} 
& \multicolumn{3}{c|}{\textbf{Design}} 
& \multicolumn{2}{c|}{\textbf{Visual-only $U$}} 
& \multicolumn{2}{c}{\textbf{Full-match $U$}} \\
\cmidrule(lr){2-4}\cmidrule(lr){5-6}\cmidrule(l){7-8}
& \textbf{Instr.} & \textbf{Transcription} & \textbf{Format} & \textbf{IoU@0.3} & \textbf{IoU@0.5} & \textbf{IoU@0.3} & \textbf{IoU@0.5} \\
\midrule
1 & Full   & Yes & XML              & \textbf{0.363} & \textbf{0.435} & \textbf{0.209} & \textbf{0.377} \\
2 & Simple & Yes & XML              & 0.276 & 0.176 & 0.062 & 0.091 \\
3 & Full   & No  & XML              & 0.341 & 0.168 & 0.173 & 0.252 \\
4 & Full   & Yes & JSON             & 0.146 & 0.254 & 0.157 & 0.223 \\
5 & Full   & Yes & Decoupled JSON   & -0.011 & -0.059 & -0.061 & -0.017 \\
6 & Full   & No  & JSON             & 0.261 & 0.200 & 0.138 & 0.273 \\
7 & Full   & No  & Decoupled JSON   & -0.017 & -0.072 & 0.107 & 0.040 \\
\bottomrule
\end{tabular}
\caption{Prompt variant sensitivity results. $U$ denotes the model-normalized utility score ($\mu - \sigma$) aggregated over pass@$k$ for $k \in \{1,2,4,8,16\}$. \textit{Visual-only} computes pass@$k$ using only the visual matching criterion, while \textit{Full-match} requires both visual and exact textual matches. All variants are evaluated on the same set of VLM backends.}
\label{tab:supp_prompt_sensitivity}
\end{table*}

\subsection{Scoring}
Let $s_{m,p,k}$ denote the pass@$k$ score for model $m$ and prompt variant $p$. We first compute the per-model effectiveness of each variant by min-max normalizing across variants and weighting by $w=(5,4,3,2,1)$ for $k \in \{1,2,4,8,16\}$:
\begin{equation}
E_{m,p} = \sum_{k} w_k \frac{s_{m,p,k} - \min_{p'} s_{m,p',k}}{\max_{p'} s_{m,p',k} - \min_{p'} s_{m,p',k}}.
\end{equation}
We then evaluate each variant across $M$ models by its average effectiveness $\mu_p = \frac{1}{M}\sum_{m=1}^M E_{m,p}$ and its robustness $\sigma_p = \sqrt{\frac{1}{M}\sum_{m=1}^M (E_{m,p} - \mu_p)^2}$. The final utility score is:
\begin{equation}
U_p = \mu_p - \sigma_p,
\end{equation}
which explicitly favors prompts that are both highly effective on average and stable across models.

\subsection{Results}
Table~\ref{tab:supp_prompt_sensitivity} reports the utility scores. Variant 1 (full instructions, transcription output, XML format) achieves the highest $U$ under both the visual-only and the stricter full-match criteria at both IoU thresholds. We therefore adopt Variant 1 as the default prompt for all single-pass evaluations in the main paper.

Two patterns emerge. First, simplifying the instructions (Variant 2) or switching to JSON output (Variants 4 and 6) reduces cross-model robustness, suggesting that G-CGEC benefits from detailed guidance and XML-style formatting. Second, the decoupled JSON prompts (Variants 5 and 7) perform worst across nearly all settings, indicating that separating correction from grounding at the output level disrupts the joint reasoning required for the task. These findings reinforce our main-paper conclusion: meaningful performance gains come from structural decomposition of the inference process (pipeline, agentic workflows) rather than prompt engineering alone.

\section{Additional Implementation Details}
\label{sec:supp_implementation_details}

\subsection{Agentic Framework and Dynamic Coordination}
Our agentic systems are built upon the OpenCode framework, which enables dynamic, on-the-fly coordination rather than relying on rigid, pre-scripted pipelines. Specifically, tasks such as merging results from sub-agents, resolving conflicts, deciding when to invoke OCR or visual grounding tools, and verifying intermediate outputs are all handled autonomously by the agents based on the context of each essay. This design allows the system to adapt its reasoning trajectory to the complexity and layout of individual inputs.

\subsection{Agentic Workflow Design}
All workflows follow an image-in, text-and-bounding-box-out protocol. We evaluate two paradigms:
\begin{itemize}
    \item \textbf{Single-Agent:} The model performs staged reasoning (reading, error detection, correction, localization, verification) and autonomously decides when to invoke OCR or visual grounding tools.
    \item \textbf{Multi-Agent:} A master agent delegates error discovery to specialized sub-agents, decomposed either by \textit{error-type} (e.g., punctuation, grammar, spelling) or by \textit{operation-type} (insertion, deletion, substitution). Sub-agents independently propose candidates, and the master agent dynamically aggregates, deduplicates, verifies against the image, and produces the final structured output.
\end{itemize}

\subsection{Pipeline String Alignment}
In the pipeline scheme, we align the OCR-transcribed text with the LLM-generated corrected text to derive the minimal sequence of insertion, deletion, and substitution edits. We use character-level Levenshtein distance for this alignment. Substitution and deletion edits inherit the bounding boxes of their aligned source spans, while insertion edits are assigned virtual character-sized boxes at the corresponding insertion anchors.

\subsection{Agent Execution Constraints}
To ensure fair comparison and prevent excessive computation from skewing results, each agentic workflow is limited to at most 30 reasoning steps per essay, and each agent run is capped at 20 minutes. If an agent exceeds either limit, the process is terminated and the last valid structured output is used for evaluation.

\subsection{Stochasticity and Reproducibility}
All experiments are conducted with temperature $T=0.3$ and top-$p=1.0$. The reported $F_1$ scores are based on a single run per configuration. Given the stochastic nature of VLM generation, minor fluctuations may occur across runs; however, the observed performance gaps between schemes are substantial and consistent.

\subsection{Tool Calling Schema}
External modules (OCR and visual grounding) are exposed to the agents as \emph{Agent Skills}. Each skill is accompanied with a SKILL.md and a Python script that can be used to call the tools. For each SKILL.md, we detail the usage of the Python script. 

\section{OCR and Visual Grounding Module Specifications}
\label{sec:supp_module_details}

\subsection{OCR Module Training and Performance}
The OCR module is trained via supervised fine-tuning (SFT) of Qwen3.5-35B-A3B on 26\,800 image essays with human-annotated transcriptions. The transcription protocol follows the same procedure described in Section~\ref{sec:dataset} of the main paper, including lenticular brackets for non-existent characters. We train for 4 epochs with a learning rate of $2 \times 10^{-5}$ and a global batch size of 64, using LoRA with rank 32 and alpha 64. The trained OCR module achieves an average Levenshtein distance of 6.84 per essay on the G-CGEC-Eval benchmark.

\subsection{Visual Grounding Module Training and Performance}
The visual grounding module is also trained on top of Qwen3.5-35B-A3B using a rejection sampling-based SFT dataset, as described in the main paper. The training data is derived from images and transcriptions of 558 essays (a subset of the OCR module's training set). For each essay, we sample up to 20 text segments of 20 characters for image index prediction and another 20 segments for character-level bounding box prediction, provided the essay is long enough. In total, the training set contains 13\,252 entries (combining image index and bounding box prediction tasks). Training hyperparameters match those of the OCR module. For evaluation, we construct a test set by sampling 20-character text segments from 50 essays in the G-CGEC-Eval benchmark and manually annotating their gold bounding boxes. Table~\ref{tab:vg_module_eval} reports the intrinsic evaluation results across different IoU thresholds.

\begin{table}[t]
\centering
\small
\setlength{\tabcolsep}{4pt}
\begin{tabular}{c c c c}
\toprule
\textbf{IoU Threshold} & \textbf{Avg. IoU} & \textbf{Recall (\%)} & \textbf{Precision (\%)} \\
\midrule
0.2 & 60.60 & 90.04 & 91.17 \\
0.5 & 67.65 & 69.00 & 69.87 \\
\bottomrule
\end{tabular}
\caption{Intrinsic evaluation results of the visual grounding module on the held-out test set. Performance is reported under varying IoU thresholds for spatial matching.}
\label{tab:vg_module_eval}
\end{table}
\end{document}

%% file: figures/error_distribution.tex
\begin{tikzpicture}
    \definecolor{cGrammar}{HTML}{4C72B0}
    \definecolor{cPunct}{HTML}{DD8452}
    \definecolor{cWrongChar}{HTML}{55A868}
    \definecolor{cLangNorms}{HTML}{C44E52}
    \definecolor{cPinyin}{HTML}{8172B2}
    \definecolor{cCharForms}{HTML}{937860}
    \definecolor{cFactual}{HTML}{DA8BC3}
    \begin{axis}[
        xbar,
        width=0.85\textwidth,
        height=8.5cm,
        bar width=2.6pt,
        bar shift=0pt,
        xmin=0, xmax=560,
        xlabel={Count (operations)},
        xlabel style={font=\scriptsize},
        ytick={0,1,2,3,4,5,6,7,8,9,10,11,12,13,14,15,16,17,18,19,20,21,22,23,24,25},
        yticklabels={
            Citation error,
            Factual error,
            Arabic numeral,
            Pinyin error,
            Unwritten pinyin,
            Subject norm,
            Aspect particle,
            Repeated subject,
            'DE' disambiguation,
            Similar shape,
            Wrong character,
            Missing punctuation,
            Sentence break,
            Wrong punctuation,
            Two-to-one mismatch,
            Ambiguity,
            Unclear reference,
            Contradiction,
            Mixed structure,
            Illogical,
            Repetition,
            Redundancy,
            Word order,
            Poor collocation,
            Inappropriate word,
            Missing component
        },
        yticklabel style={font=\scriptsize, anchor=east},
        tick label style={font=\scriptsize},
        axis y line*=left,
        axis x line*=bottom,
        axis line style={line width=0.5pt},
        tick style={line width=0.5pt},
        tick align=outside,
        nodes near coords,
        nodes near coords style={font=\scriptsize},
        every node near coord/.append style={xshift=2pt},
        xmajorgrids=false,
        ymin=-0.5, ymax=25.5,
        clip=false,
        xtick pos=left,
    ]

    \addplot[fill=cGrammar, draw=none] coordinates {
        (1,14) (2,15) (2,16) (8,17) (32,18) (82,19) (102,20)
        (134,21) (145,22) (169,23) (342,24) (385,25)
    };

    \addplot[fill=cPunct, draw=none] coordinates {
        (223,11) (305,12) (477,13)
    };

    \addplot[fill=cWrongChar, draw=none] coordinates {
        (201,9) (317,10)
    };

    \addplot[fill=cLangNorms, draw=none] coordinates {
        (3,5) (22,6) (33,7) (119,8)
    };

    \addplot[fill=cPinyin, draw=none] coordinates {
        (4,3) (35,4)
    };

    \addplot[fill=cCharForms, draw=none] coordinates {
        (29,2)
    };

    \addplot[fill=cFactual, draw=none] coordinates {
        (1,0) (15,1)
    };

    \draw[gray, line width=0.5pt] (axis cs:0,1.5) -- (axis cs:560,1.5);
    \draw[gray, line width=0.5pt] (axis cs:0,2.5) -- (axis cs:560,2.5);
    \draw[gray, line width=0.5pt] (axis cs:0,4.5) -- (axis cs:560,4.5);
    \draw[gray, line width=0.5pt] (axis cs:0,8.5) -- (axis cs:560,8.5);
    \draw[gray, line width=0.5pt] (axis cs:0,10.5) -- (axis cs:560,10.5);
    \draw[gray, line width=0.5pt] (axis cs:0,13.5) -- (axis cs:560,13.5);

    \node[anchor=west, text=cFactual, font=\bfseries\scriptsize] at (axis cs:400,0.5) {Factual (16)};
    \node[anchor=west, text=cCharForms, font=\bfseries\scriptsize] at (axis cs:400,2) {Character Forms (29)};
    \node[anchor=west, text=cPinyin, font=\bfseries\scriptsize] at (axis cs:400,3.5) {Pinyin (39)};
    \node[anchor=west, text=cLangNorms, font=\bfseries\scriptsize] at (axis cs:400,6.5) {Language Norms (177)};
    \node[anchor=west, text=cWrongChar, font=\bfseries\scriptsize] at (axis cs:400,9.5) {Wrong Character (518)};
    \node[anchor=west, text=cPunct, font=\bfseries\scriptsize] at (axis cs:400,12) {Punctuation (1005)};
    \node[anchor=west, text=cGrammar, font=\bfseries\scriptsize] at (axis cs:400,19.5) {Syntax \& Discourse (1404)};

    \end{axis}
    \end{tikzpicture}

%% file: figures/failure_decomposition_counts.tex
\begin{tikzpicture}
    \pgfplotsset{
      every axis/.append style={font=\small},
    }
    \definecolor{cTP}{RGB}{34,139,34}
    \definecolor{cFPloc}{RGB}{230,159,0}
    \definecolor{cFPboth}{RGB}{213,94,0}
    \definecolor{cFPc}{RGB}{190,20,20}
    \definecolor{cFNloc}{RGB}{203,201,226}
    \definecolor{cFNboth}{RGB}{158,154,200}
    \definecolor{cFNc}{RGB}{123,80,168}
    \tikzset{
      pctFPloc/.style={font=\small, text=white, fill=cFPloc, inner sep=0.6pt},
      pctFPboth/.style={font=\small, text=white, fill=cFPboth, inner sep=0.6pt},
      pctFPc/.style={font=\small, text=white, fill=cFPc, inner sep=0.6pt},
      pctFNloc/.style={font=\small, text=white, fill=cFNloc, inner sep=0.6pt},
      pctFNboth/.style={font=\small, text=white, fill=cFNboth, inner sep=0.6pt},
      pctFNc/.style={font=\small, text=white, fill=cFNc, inner sep=0.6pt},
    }
    \begin{axis}[
        width=0.95\textwidth,
        height=6.35cm,
        bar width=11pt,
        ymin=0, ymax=3080,
        ytick={0,500,1000,1500,2000,2500},
        ylabel={Count},
        xmin=-0.7, xmax=17.7,
        xtick={1,4,7,10,13,16},
        xticklabels={%
          {GPT-5.5\\Agentic},%
          {GPT-5.5\\Pipeline},%
          {MiMo-V2.5\\Agentic},%
          {MiMo-V2.5\\Pipeline},%
          {Qwen3.5-27B\\Agentic},%
          {Qwen3.5-27B\\Pipeline}},
        xticklabel style={align=center, anchor=north, font=\small},
        ymajorgrids=true,
        grid style={dotted, gray!45, line width=0.5pt},
        axis line style={line width=0.5pt},
        tick style={line width=0.5pt},
        legend style={
          at={(0.5,1.04)}, anchor=south, legend columns=7,
          draw=none, fill=none, font=\small,
          /tikz/every even column/.append style={column sep=5pt},
        },
        legend image code/.code={\draw[fill=#1,draw=none](0cm,-0.08cm) rectangle (0.24cm,0.16cm);},
    ]


    \addlegendimage{area legend, fill=cFPloc, draw=none}\addlegendentry{FP: visual}
    \addlegendimage{area legend, fill=cFPboth, draw=none}\addlegendentry{FP: both}
    \addlegendimage{area legend, fill=cFPc, draw=none}\addlegendentry{FP: textual}
    \addlegendimage{area legend, fill=cTP, draw=none}\addlegendentry{TP}
    \addlegendimage{area legend, fill=cFNloc, draw=none}\addlegendentry{FN: visual}
    \addlegendimage{area legend, fill=cFNboth, draw=none}\addlegendentry{FN: both}
    \addlegendimage{area legend, fill=cFNc, draw=none}\addlegendentry{FN: textual}

    \draw[fill=cTP, draw=none] (axis cs:-0.35,0) rectangle (axis cs:0.55,504);
    \draw[fill=cTP, draw=none] (axis cs:2.65,0) rectangle (axis cs:3.55,288);
    \draw[fill=cTP, draw=none] (axis cs:5.65,0) rectangle (axis cs:6.55,359);
    \draw[fill=cTP, draw=none] (axis cs:8.65,0) rectangle (axis cs:9.55,311);
    \draw[fill=cTP, draw=none] (axis cs:11.65,0) rectangle (axis cs:12.55,96);
    \draw[fill=cTP, draw=none] (axis cs:14.65,0) rectangle (axis cs:15.55,305);

    \draw[fill=cFPloc, draw=none] (axis cs:0.55,0) rectangle (axis cs:1.45,254);
    \draw[fill=cFPloc, draw=none] (axis cs:3.55,0) rectangle (axis cs:4.45,222);
    \draw[fill=cFPloc, draw=none] (axis cs:6.55,0) rectangle (axis cs:7.45,160);
    \draw[fill=cFPloc, draw=none] (axis cs:9.55,0) rectangle (axis cs:10.45,210);
    \draw[fill=cFPloc, draw=none] (axis cs:12.55,0) rectangle (axis cs:13.45,106);
    \draw[fill=cFPloc, draw=none] (axis cs:15.55,0) rectangle (axis cs:16.45,208);

    \draw[fill=cFPboth, draw=none] (axis cs:0.55,254) rectangle (axis cs:1.45,839);
    \draw[fill=cFPboth, draw=none] (axis cs:3.55,222) rectangle (axis cs:4.45,1234);
    \draw[fill=cFPboth, draw=none] (axis cs:6.55,160) rectangle (axis cs:7.45,567);
    \draw[fill=cFPboth, draw=none] (axis cs:9.55,210) rectangle (axis cs:10.45,1017);
    \draw[fill=cFPboth, draw=none] (axis cs:12.55,106) rectangle (axis cs:13.45,618);
    \draw[fill=cFPboth, draw=none] (axis cs:15.55,208) rectangle (axis cs:16.45,1124);

    \draw[fill=cFPc, draw=none] (axis cs:0.55,839) rectangle (axis cs:1.45,1018);
    \draw[fill=cFPc, draw=none] (axis cs:3.55,1234) rectangle (axis cs:4.45,1477);
    \draw[fill=cFPc, draw=none] (axis cs:6.55,567) rectangle (axis cs:7.45,700);
    \draw[fill=cFPc, draw=none] (axis cs:9.55,1017) rectangle (axis cs:10.45,1236);
    \draw[fill=cFPc, draw=none] (axis cs:12.55,618) rectangle (axis cs:13.45,708);
    \draw[fill=cFPc, draw=none] (axis cs:15.55,1124) rectangle (axis cs:16.45,1327);

    \draw[fill=cFNloc, draw=none] (axis cs:1.45,0) rectangle (axis cs:2.35,254);
    \draw[fill=cFNboth, draw=none] (axis cs:1.45,254) rectangle (axis cs:2.35,2210);
    \draw[fill=cFNc, draw=none] (axis cs:1.45,2210) rectangle (axis cs:2.35,2389);
    \draw[fill=cFNloc, draw=none] (axis cs:4.45,0) rectangle (axis cs:5.35,222);
    \draw[fill=cFNboth, draw=none] (axis cs:4.45,222) rectangle (axis cs:5.35,2359);
    \draw[fill=cFNc, draw=none] (axis cs:4.45,2359) rectangle (axis cs:5.35,2602);
    \draw[fill=cFNloc, draw=none] (axis cs:7.45,0) rectangle (axis cs:8.35,160);
    \draw[fill=cFNboth, draw=none] (axis cs:7.45,160) rectangle (axis cs:8.35,2401);
    \draw[fill=cFNc, draw=none] (axis cs:7.45,2401) rectangle (axis cs:8.35,2534);
    \draw[fill=cFNloc, draw=none] (axis cs:10.45,0) rectangle (axis cs:11.35,210);
    \draw[fill=cFNboth, draw=none] (axis cs:10.45,210) rectangle (axis cs:11.35,2362);
    \draw[fill=cFNc, draw=none] (axis cs:10.45,2362) rectangle (axis cs:11.35,2581);
    \draw[fill=cFNloc, draw=none] (axis cs:13.45,0) rectangle (axis cs:14.35,106);
    \draw[fill=cFNboth, draw=none] (axis cs:13.45,106) rectangle (axis cs:14.35,2701);
    \draw[fill=cFNc, draw=none] (axis cs:13.45,2701) rectangle (axis cs:14.35,2791);
    \draw[fill=cFNloc, draw=none] (axis cs:16.45,0) rectangle (axis cs:17.35,208);
    \draw[fill=cFNboth, draw=none] (axis cs:16.45,208) rectangle (axis cs:17.35,2384);
    \draw[fill=cFNc, draw=none] (axis cs:16.45,2384) rectangle (axis cs:17.35,2587);

    \node[font=\small, anchor=south] at (axis cs:0.1,504) {504};
    \node[font=\small, anchor=south] at (axis cs:3.1,288) {288};
    \node[font=\small, anchor=south] at (axis cs:6.1,359) {359};
    \node[font=\small, anchor=south] at (axis cs:9.1,311) {311};
    \node[font=\small, anchor=south] at (axis cs:12.1,96) {96};
    \node[font=\small, anchor=south] at (axis cs:15.1,305) {305};
    \node[font=\small, anchor=south] at (axis cs:1,1018) {1018};
    \node[font=\small, anchor=south] at (axis cs:4,1477) {1477};
    \node[font=\small, anchor=south] at (axis cs:7,700) {700};
    \node[font=\small, anchor=south] at (axis cs:10,1236) {1236};
    \node[font=\small, anchor=south] at (axis cs:13,708) {708};
    \node[font=\small, anchor=south] at (axis cs:16,1327) {1327};
    \node[font=\small, anchor=south] at (axis cs:1.9,2389) {2389};
    \node[font=\small, anchor=south] at (axis cs:4.9,2602) {2602};
    \node[font=\small, anchor=south] at (axis cs:7.9,2534) {2534};
    \node[font=\small, anchor=south] at (axis cs:10.9,2581) {2581};
    \node[font=\small, anchor=south] at (axis cs:13.9,2791) {2791};
    \node[font=\small, anchor=south] at (axis cs:16.9,2587) {2587};

    \node[pctFPloc] at (axis cs:1,127) {25\%};
    \node[pctFPboth] at (axis cs:1,546) {57\%};
    \node[pctFPc] at (axis cs:1,928) {18\%};
    \node[pctFPloc] at (axis cs:4,111) {15\%};
    \node[pctFPboth] at (axis cs:4,728) {69\%};
    \node[pctFPc] at (axis cs:4,1355) {16\%};
    \node[pctFPloc] at (axis cs:7,80) {23\%};
    \node[pctFPboth] at (axis cs:7,363) {58\%};
    \node[pctFPc] at (axis cs:7,633) {19\%};
    \node[pctFPloc] at (axis cs:10,105) {17\%};
    \node[pctFPboth] at (axis cs:10,613) {65\%};
    \node[pctFPc] at (axis cs:10,1126) {18\%};
    \node[pctFPloc] at (axis cs:13,53) {15\%};
    \node[pctFPboth] at (axis cs:13,362) {72\%};
    \node[pctFPc] at (axis cs:13,663) {13\%};
    \node[pctFPloc] at (axis cs:16,104) {16\%};
    \node[pctFPboth] at (axis cs:16,666) {69\%};
    \node[pctFPc] at (axis cs:16,1225) {15\%};

    \node[pctFNloc] at (axis cs:1.9,127) {11\%};
    \node[pctFNboth] at (axis cs:1.9,1232) {82\%};
    \node[pctFNc] at (axis cs:1.9,2299) {7\%};
    \node[pctFNloc] at (axis cs:4.9,111) {9\%};
    \node[pctFNboth] at (axis cs:4.9,1290) {82\%};
    \node[pctFNc] at (axis cs:4.9,2480) {9\%};
    \node[pctFNloc] at (axis cs:7.9,80) {6\%};
    \node[pctFNboth] at (axis cs:7.9,1280) {89\%};
    \node[pctFNc] at (axis cs:7.9,2467) {5\%};
    \node[pctFNloc] at (axis cs:10.9,105) {8\%};
    \node[pctFNboth] at (axis cs:10.9,1286) {83\%};
    \node[pctFNc] at (axis cs:10.9,2471) {9\%};
    \node[pctFNloc] at (axis cs:13.9,53) {4\%};
    \node[pctFNboth] at (axis cs:13.9,1403) {93\%};
    \node[pctFNc] at (axis cs:13.9,2746) {3\%};
    \node[pctFNloc] at (axis cs:16.9,104) {8\%};
    \node[pctFNboth] at (axis cs:16.9,1296) {84\%};
    \node[pctFNc] at (axis cs:16.9,2485) {8\%};

    \end{axis}
\end{tikzpicture}

%% file: figures/center_distance_curve.tex
\begin{tikzpicture}
    \definecolor{singleblue}{RGB}{31,119,180}
    \definecolor{pipelineorange}{RGB}{255,127,14}
    \definecolor{agenticgreen}{RGB}{44,160,44}
    \begin{axis}[
        width=\linewidth,
        height=0.50\linewidth,
        xmin=-0.15,
        xmax=6.15,
        ymin=-1,
        ymax=36.5,
        xtick={0,1,2,3,4,5,6},
        xticklabels={18,38,80,160,320,640,1000},
        enlarge x limits=false,
        ytick={0,5,10,15,20,25,30,35},
        xlabel={Center-distance (CD) threshold (log-scale)},
        ylabel={$F_1$},
        xlabel style={font=\small, at={(axis description cs:0.5,-0.20)}, anchor=north},
        ylabel style={font=\small},
        tick label style={font=\small},
        grid=major,
        grid style={densely dotted, gray!35, line width=0.5pt},
        axis line style={line width=0.5pt},
        tick style={line width=0.5pt},
        clip=false,
        legend style={
            at={(0.5,-0.39)},
            anchor=north,
            legend columns=3,
            font=\small,
            draw=none,
            /tikz/every even column/.append style={column sep=1.5mm},
            row sep=0pt,
        },
    ]
        \addlegendimage{color=singleblue, line width=0.75pt}
        \addlegendentry{Single-pass}
        \addlegendimage{color=pipelineorange, line width=0.75pt}
        \addlegendentry{Pipeline}
        \addlegendimage{color=agenticgreen, line width=0.75pt}
        \addlegendentry{Agentic}

        \addlegendimage{black, only marks, mark=square*, mark size=1.55pt, line width=0.75pt}
        \addlegendentry{GPT-5.5}
        \addlegendimage{black, only marks, mark=triangle*, mark size=1.75pt, line width=0.75pt}
        \addlegendentry{MiMo-V2.5}
        \addlegendimage{black, only marks, mark=*, mark size=1.55pt, line width=0.75pt}
        \addlegendentry{Qwen3.5-27B}

        \addplot[color=agenticgreen, line width=0.75pt, mark=square*, mark size=1.55pt, solid, forget plot]
            coordinates {(0,25.142) (1,29.626)};
        \addplot[color=agenticgreen, line width=0.75pt, mark=square*, mark size=1.55pt, dashed, forget plot]
            coordinates {(1,29.626) (2,31.302) (3,32.292) (4,33.560) (5,34.601) (6,34.828)};

        \addplot[color=agenticgreen, line width=0.75pt, mark=triangle*, mark size=1.75pt, solid, forget plot]
            coordinates {(0,19.484) (1,21.711)};
        \addplot[color=agenticgreen, line width=0.75pt, mark=triangle*, mark size=1.75pt, dashed, forget plot]
            coordinates {(1,21.711) (2,23.229) (3,24.443) (4,25.436) (5,26.903) (6,27.307)};

        \addplot[color=agenticgreen, line width=0.75pt, mark=*, mark size=1.55pt, solid, forget plot]
            coordinates {(0,5.797) (1,6.876)};
        \addplot[color=agenticgreen, line width=0.75pt, mark=*, mark size=1.55pt, dashed, forget plot]
            coordinates {(1,6.876) (2,8.287) (3,9.049) (4,10.132) (5,11.111) (6,11.321)};

        \addplot[color=pipelineorange, line width=0.75pt, mark=square*, mark size=1.55pt, solid, forget plot]
            coordinates {(0,13.664) (1,17.564)};
        \addplot[color=pipelineorange, line width=0.75pt, mark=square*, mark size=1.55pt, dashed, forget plot]
            coordinates {(1,17.564) (2,19.322) (3,20.180) (4,21.201) (5,22.486) (6,22.690)};

        \addplot[color=pipelineorange, line width=0.75pt, mark=triangle*, mark size=1.75pt, solid, forget plot]
            coordinates {(0,15.495) (1,19.324)};
        \addplot[color=pipelineorange, line width=0.75pt, mark=triangle*, mark size=1.75pt, dashed, forget plot]
            coordinates {(1,19.324) (2,21.031) (3,21.652) (4,22.462) (5,23.498) (6,23.723)};

        \addplot[color=pipelineorange, line width=0.75pt, mark=*, mark size=1.55pt, solid, forget plot]
            coordinates {(0,14.898) (1,18.664)};
        \addplot[color=pipelineorange, line width=0.75pt, mark=*, mark size=1.55pt, dashed, forget plot]
            coordinates {(1,18.664) (2,19.947) (3,20.787) (4,21.274) (5,22.822) (6,23.043)};

        \addplot[color=singleblue, line width=0.75pt, mark=square*, mark size=1.55pt, solid, forget plot]
            coordinates {(0,0.380) (1,1.248)};
        \addplot[color=singleblue, line width=0.75pt, mark=square*, mark size=1.55pt, dashed, forget plot]
            coordinates {(1,1.248) (2,3.267) (3,9.298) (4,16.359) (5,20.601) (6,21.410)};

        \addplot[color=singleblue, line width=0.75pt, mark=triangle*, mark size=1.75pt, solid, forget plot]
            coordinates {(0,0.062) (1,0.431)};
        \addplot[color=singleblue, line width=0.75pt, mark=triangle*, mark size=1.75pt, dashed, forget plot]
            coordinates {(1,0.431) (2,1.480) (3,3.437) (4,6.024) (5,7.560) (6,7.985)};

        \addplot[color=singleblue, line width=0.75pt, mark=*, mark size=1.55pt, solid, forget plot]
            coordinates {(0,1.288) (1,2.372)};
        \addplot[color=singleblue, line width=0.75pt, mark=*, mark size=1.55pt, dashed, forget plot]
            coordinates {(1,2.372) (2,4.520) (3,7.534) (4,11.616) (5,14.499) (6,15.096)};
    \end{axis}
    \end{tikzpicture}